\documentclass[pdflatex,sn-mathphys-ay]{sn-jnl}

\usepackage{amsmath,amssymb,amsfonts}%
\usepackage{amsthm}%
\usepackage{mathrsfs}%

\usepackage{siunitx}
\usepackage{array}
\usepackage{color}
\usepackage{tikz}
\usepackage{pgfgantt}
\usepackage{url}
\usepackage{makecell}
\usepackage{caption}
\usepackage{subcaption}
\usepackage{multirow}
\usepackage{graphicx}%
\usepackage{multirow}%
\usepackage{amsthm}%
\usepackage{mathrsfs}%
\usepackage[title]{appendix}%
\usepackage{xcolor}%
\usepackage{textcomp}%
\usepackage{manyfoot}%
\usepackage{booktabs}%
\usepackage{algorithm}%
\usepackage{algorithmicx}%
\usepackage{algpseudocode}%
\usepackage{listings}%
\usepackage{nicematrix}
\usepackage{anyfontsize}

\begin{document}

\title[Article Title]{Deep Learning Techniques for Atmospheric Turbulence Removal: A Review}
\author*[1]{\fnm{Paul} \sur{Hill}}\email{paul.hill@bristol.ac.uk}
\author[1]{\fnm{Nantheera} \sur{Anantrasirichai}}
\author[1]{\fnm{Alin} \sur{Achim}}
\author[1]{\fnm{David} \sur{Bull}}
\affil*[1]{\orgname{The University of Bristol}}

\abstract{\noindent The influence of atmospheric turbulence on acquired imagery makes image interpretation and scene analysis extremely difficult and reduces the effectiveness of conventional approaches for classifying and tracking objects of interest in the scene. Restoring a scene distorted by
atmospheric turbulence is also a challenging problem. The effect, which is caused by random,
spatially varying perturbations, makes conventional model-based approaches difficult and, in most cases,
impractical due to complexity and memory requirements. Deep learning approaches offer faster operation and are capable of implementation on small devices. This paper reviews the characteristics of atmospheric turbulence and its impact on acquired imagery. It compares the performance of various state-of-the-art deep neural networks, including Transformers, SWIN and Mamba,  when used to mitigate spatio-temporal image distortions.}


\maketitle

\section{Introduction}
\label{sec:intro}

The imaging of distant objects within the atmosphere is subject to, and often limited by, atmospheric turbulence.  Atmospheric turbulence degrades visual quality and impacts the performance of automated recognition and tracking of objects of interest within the scene. Such distortions occur when the temperature difference between the ground and the air increases, causing the air layers to move upwards rapidly, leading to spatially varying changes in the index of refraction along the optical path.  This is generally observed as a combination of blur, ripple, and intensity fluctuations in the scene. Examples of this effect are found at locations such as hot roads and deserts, as well as in the proximity of hot human-made objects such as aircraft jet exhausts. This effect is most noticeable close to the ground in hot environments and can combine with other detrimental effects in long-range image acquisition (e.g.\ surveillance) applications, such as fog or haze, which reduce contrast and video quality.  Significant other factors include altitude, wind speed, humidity, and pollution.  Perfect mitigation of atmospheric turbulence effects is an ill-posed problem due to the spatio-temporally varying nature of the distortions. 

\begin{figure}
    \centering
    \includegraphics[width=0.55\textwidth]{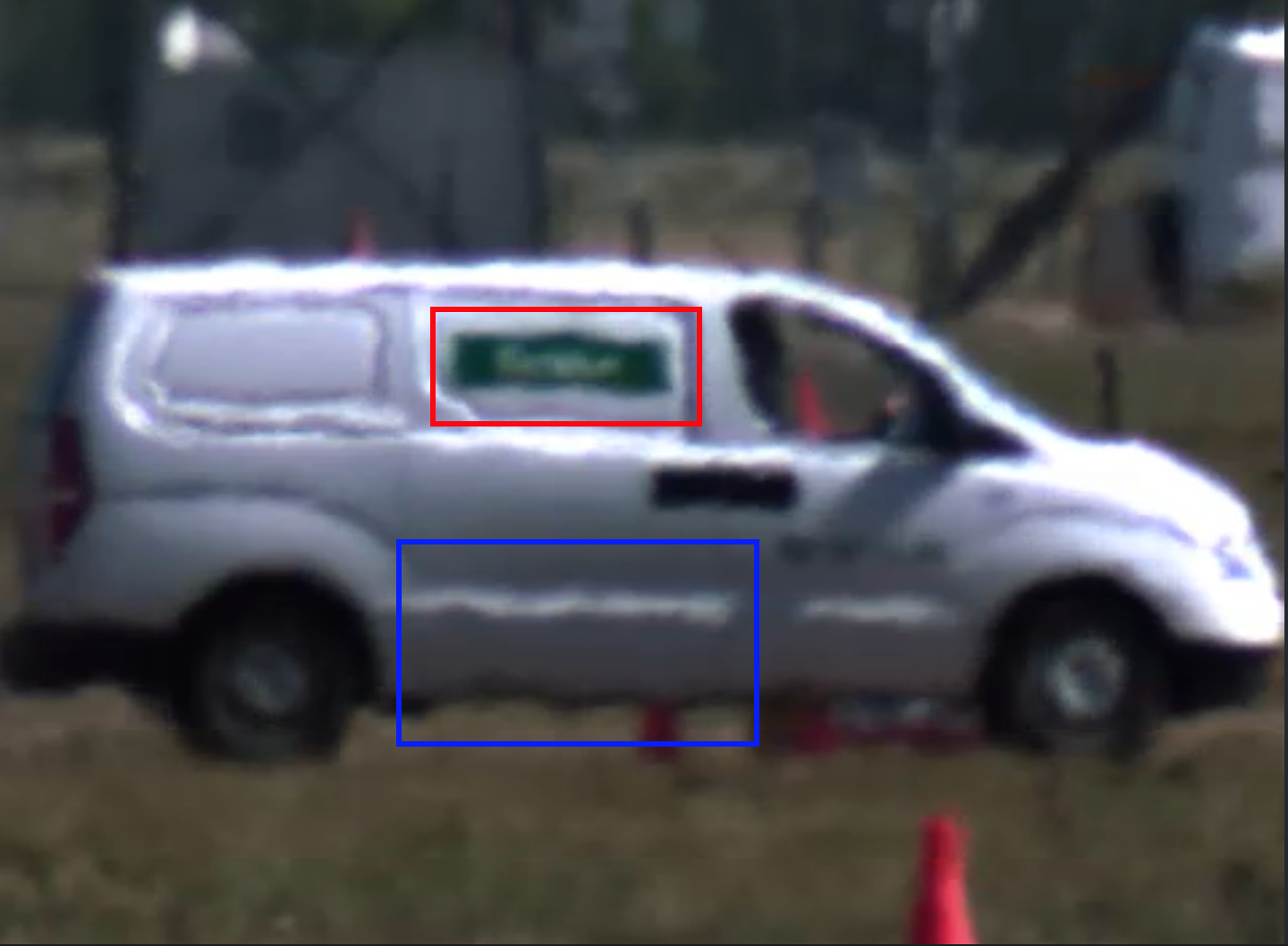}
    \caption[]{Atmospheric distortions:  Frame from the real atmospheric distortion ``Van'' sequence.   The blue rectangle illustrates pixel displacements resulting in non-linear lines and the red rectangle illustrates the spatially varying blur effect.}
    \label{fig:distortion}
\end{figure}

\subsection{Atmospheric Turbulence Visual Effects}
Figure \ref{fig:distortion} illustrates the main two visual effects of atmospheric turbulence: spatially varying pixel displacements and blurring. The pixel displacements (sometimes referred to as ``tilt'' or ``dancing'') result in spatial distortions;  the most easily recognisable visual feature being that straight lines become ``wavy''.  Within the real datasets available (see section~\ref{sec:datasets}), there is no large observable effect of amplitude variations i.e.\ turbulence does not significantly vary the brightness/contrast of the imaging output spatially or over time (although often, environmental factors such as haze degrades the contrast of the image.)  Therefore it can be assumed the 
 degrading effect is caused by phase distortions.

\subsection{Atmospheric Turbulence Effect Removal}
\begin{figure}
    \centering
    \includegraphics[width=0.4\textwidth]{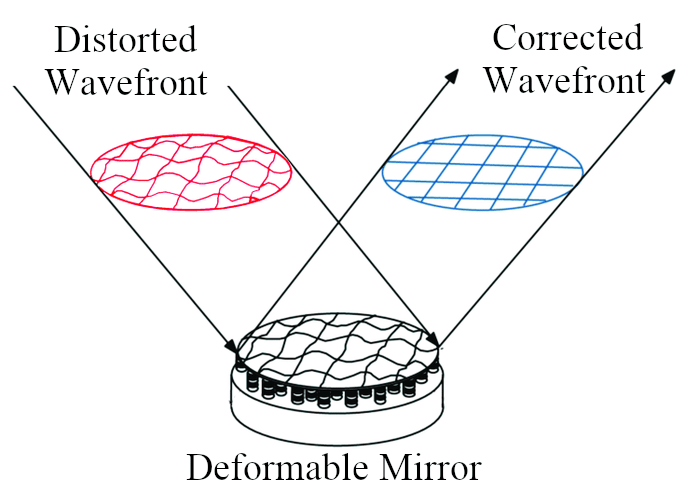}
    \caption[Caption for LOF]{Deformable mirror for adaptive optics based turbulence mitigation: adapted from~[\cite{SteinbockMSc:2012}].}
    \label{fig:def_optics}
\end{figure}

Atmospheric turbulence removal can be categorised into two types of methods: adaptive optics based methods [\cite{Roggemann:Imaging:1996,TysonBook:2015,Pearson:1976}] and image processing based methods [\cite{Anantrasirichai:Atmospheric:2013, Zhu:Removing:2013, Xue:Video:2016, Patel:adaptive:2019, Deledalle:blind:2020,Oreifej:Simultaneous:2013,Chen:detecting:2014, Halder:geometric:2015, Zhang:Stabilization;2018,Elahi:Detecting:2018,Foi:methods:2015, Anantrasirichai:Atmospheric:2018, Nieuwenhuizen:Dynamic:2019, Mao:Image:2020, Boehrer:Turbulence:2021,Vint:analysis:2020,Chak:Subsampled:2021,Mao:accelaring:2021,Wang:deep:2021,Cheng:Restoration:2023,Li:Unsupervised:2021,Gao:Atmospheric:2019, chan2023computational}].  Techniques using adaptive optics require complex and expensive
hardware.  An example of an adaptive optical system is shown in Figure \ref{fig:def_optics} where a deformable mirror is used to reform a coherent wavefront within an optical capture system.  Adaptive optics also do not solve all optical effects of atmospheric turbulence, as the amount of turbulence depends on temperature variation.  This paper therefore focuses on image processing-based mitigation methods.

 Modern deep learning has emerged as a powerful tool to find patterns, analyse information, and predict future events [\cite{Anantrasirichai:Artificial:2021,Goodfellow-et-al-2016}.
\cite{Vint:analysis:2020}] investigated the performance of supervised deep learning for turbulence mitigation in long-range imagery comparing various state-of-the-art architectures, originally proposed for denoising, segmentation, and super-resolution. 
This study however only tested Convolutional Neural Networks (CNNs) using one synthetic static scene. In contrast, in this paper, we consider a variety of real situations where moving objects are distorted by atmospheric turbulence.

\subsection{Paper Structure}

In this paper, we characterise the impact of atmospheric distortions on acquired imagery and provide a review of mitigation from both a historical and state-of-the-art perspective. In particular, we assess the performance of deep learning methods in removing these distortions.

The remainder of this paper is organised as follows: Section \ref{sec:Characteristics} lists the properties of atmospheric turbulence and its impact on imaging systems. This is valuable to better understand the problem and as a basis for generating synthetic distortions.
We first briefly describe traditional model-based approaches where the nature of the distortions is studied and a mathematical model formed to describe it.
Training data is crucial for data-driven methods like deep learning, and a wide range of datasets (both real and artificial) are described in section \ref{sec:datasets} together with a review of existing methods for generating artificial turbulence sequences.
Conventional model based methods for turbulence mitigation are reviewed in section \ref{sec:turbModels}.  For deep learning techniques, we investigate state-of-the-art approaches in section \ref{sec:deep} proposed specifically for atmospheric turbulence removal and for restoration tasks. These tasks include denoising (as atmospheric turbulence distortion can be considered as noise), deblurring (as temporal averaging not only reduces geometric distortion but creates blur) and super-resolution (as details of the distorted scene need to be enhanced).  In particularly we present results for a modified version of an effective learning-based method,TMT [\cite{zhang2024imaging}] by changing convolutional layers to transformer based SWIN3D~[\cite{yang2023swin3d}] and the state space model, Mamba3D~[\cite{gong2024nnmamba}].
Finally, in section \ref{sec:modelA} we analyse a range of architectures and their hyper-parameters for some simple models,  using feature-map energy to ascertain what parts of the network are important for various conditions and sequences.

\section{Characterising Atmospheric Turbulence}
\label{sec:Characteristics}

When the temperature variation between the ground and the air increases, the thickness of each layer decreases and the air layers move upwards rapidly, leading to faster and greater
micro-scale changes in the air’s refractive index. The strength of atmospheric turbulence is commonly described by the refractive-index structure constant, $C_n^2$, where 0 represents turbulence-free. $C_n^2$ of the air typically ranges from 10$^{-17}$ to 10$^{-13}$ m$^{-2/3}$, while $C_n^2$ typically increases by a factor of eight in summer daytime~[\cite{Schwartzman:Turbulence:2017}]. The light propagates through this inhomogeneous refractive-index medium resulting in ripples or wavy effects in the acquired video. These turbulent motions have the following characteristics: 

\begin{itemize}
    \item \textbf{Fluctuations in velocity}. This can be explained using the Reynolds decomposition: $u(t) = \bar{u} + u'(t)$, where $\bar{u}$ is the mean of velocity $u$. $u'(t)$ is the time varying turbulent fluctuation (the time average of which equals zero~[\cite{Adrian:Analysis:2000}]). This effect, commonly referred to as ``image dancing'' affects local image displacements between consecutive frames rather than global shifts only~[\cite{Huebner:compensating:2009}]. 
    \item \textbf{Quasi-periodicity}: Atmospheric turbulence is quasi-periodic~[\cite{Li:Suppressing:2009}], meaning that the ripple has an unpredictable pattern of recurrence. Therefore, averaging a number of frames would result in geometrical improvement, but the result will be blurred by an unknown Point Spread Function (PSF) of the same size as the pixel motions due to the turbulence.
    \item \textbf{Chaotic process}: A small change in the initial conditions of fluid velocity $u(x, t_0)$ results in a large change in the conditions at a later time $t$~[\cite{Ryden:Turbulence:2011}]. Atmospheric turbulence is thus unpredictable in space and time.
    \item \textbf{Anisoplanatism}: Capturing sequences through hot-air turbulence generally results in mixed signals manifesting as wavy ripples in the large distorted areas and as ``jittering'' in the smaller distorted ones~[\cite{Wu:Method:2018}]. Image ``dancing'' (displacements within the image plane) will affect local image displacements between consecutive frames rather than global shifts only~[\cite{Huebner:compensating:2009}]. The mixing signals cause image aberration and the longer optical path length and stronger turbulence cause more severe aberrations and reduced isoplanatic angle.
\end{itemize}

\noindent During image acquisition, light rays reflecting off objects travelling through atmospheric turbulence experience different delays, leading to wave interference and phase diversity. This results in local tilt, blur, and geometric distortion. Previous work has modelled these effects in the phase domain (e.g.~[\cite{Woods:Lucky:2009,Grabner:Atmospheric:2011,Anantrasirichai:deep:2019,Mao:accelaring:2021}]), reduce displacement in the phase domain~[\cite{Anantrasirichai:Atmospheric:2013}] and generate the synthetic data in the phase domain (see Section \ref{ssec:syndata}). Other works model the distortion directly in the image domain (e.g.\ using intensity, motion and PSFs~[\cite{Zhu:Removing:2013, Oreifej:Simultaneous:2013, Patel:adaptive:2019}]).

\section{Datasets}
\label{sec:datasets}

Turbulence datasets are key to the development and evaluation of learning-based solutions. These can be categorised as real or synthetic.  Synthetic data adds artificial turbulence to clean imagery therefore leading to a dataset containing ground truth and distorted versions.  Conversely, real turbulence data may or may not have ground truth data.  Pseudo ground truth for real turbulence data is generated as described below. 

This section also lists and describes available datasets.  Many more datasets have been used in the literature.  However, we list here only datasets publicly available.  This aims to help researchers in the field to model their work and to test it with various levels of image distortions. The synthetic datasets, where the properties of the atmospheric turbulence described in the previous section are employed to create the distortions, are generally less severe than real datasets, but the ground truth is available. It can also be generated as many times as needed. However, the simplicity of the models could limit the performance when testing with real sequences, as the models may overfit to the dataset.  A representative subset of the most popular data available in the literature is shown in Table 1.

\begin{table}[!ht]
    \caption{Representative real turbulence data with and without ground truth}
    \begin{tabular}{m{0.06\linewidth}m{0.22\linewidth}m{0.06\linewidth}m{0.15\linewidth}m{0.3\linewidth}}
    \hline
    Thumb&Sequence&Frames&Resolution&Content\\ 
    \hline
    \includegraphics[width=0.07\textwidth]{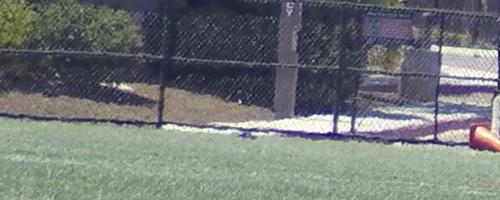}&OTIS-Car~\cite{Gilles:Open:2017}&100&\makecell[l]{Various \\ (e.g. 237$\times$237)}&Moving small toy car, Outdoor, Static/Dynamic, Real, GT, Low Turbulence\\
    \includegraphics[width=0.07\textwidth]{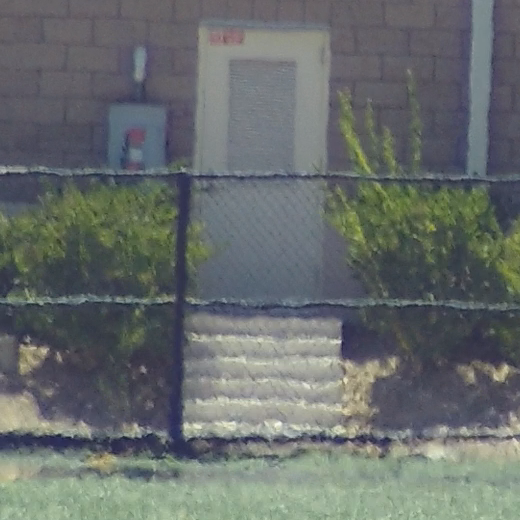}&OTIS-Door~\cite{Gilles:Open:2017}&300&520$\times$520&Door, Outdoor, Static, Real, No-GT, Medium Turbulence\\
    \includegraphics[width=0.07\textwidth]{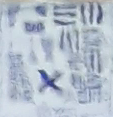}&OTIS-Pattern~\cite{Gilles:Open:2017}&100&\makecell[l]{Various \\(e.g.\ 113$\times$117)}&USAF Pattern, Indoor, Static, Real, GT, High Turbulence\\
    \includegraphics[width=0.07\textwidth]{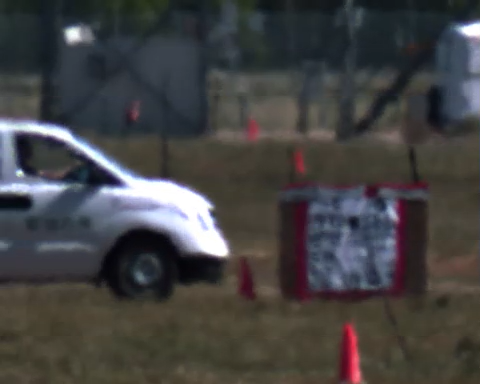}&Van~\cite{CLEAR:data2}&159&480$\times$384&Moving van\&USAF chart, Outdoor, Dynamic, Real, No-GT, Medium Turbulence\\
    \includegraphics[width=0.07\textwidth]{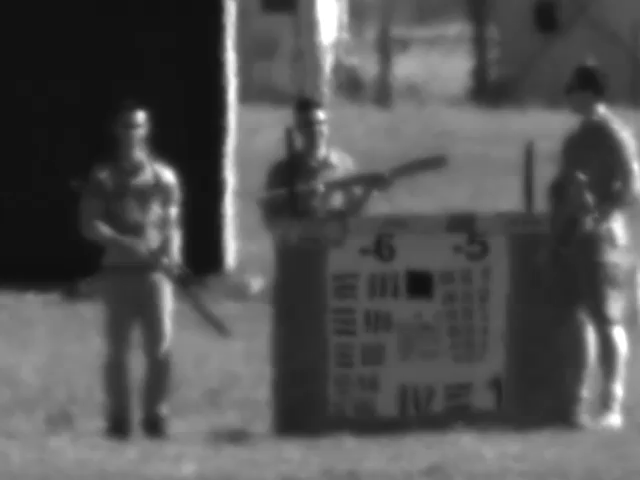}&Men~\cite{CLEAR:data2}&100&640$\times$480&Three men\&USAF chart, Outdoor, Static, Real, No-GT, Medium Turbulence\\
    \includegraphics[width=0.07\textwidth]{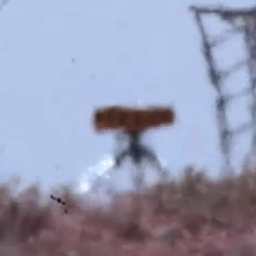}&Numberplate~\cite{CLEAR:data2}&264&256$\times$256&Number plate, Outdoor, Static, Real, No-GT, High Turbulence\\
    \includegraphics[width=0.07\textwidth]{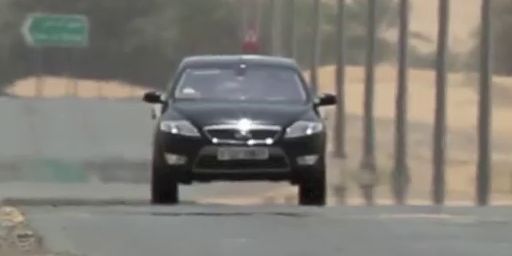}&Car~\cite{CLEAR:data}&163&512$\times$256&Single Moving Car, Outdoor (distance$\sim$500m), Dynamic, Real, No-GT, Medium Turbulence\\
    \includegraphics[width=0.07\textwidth]{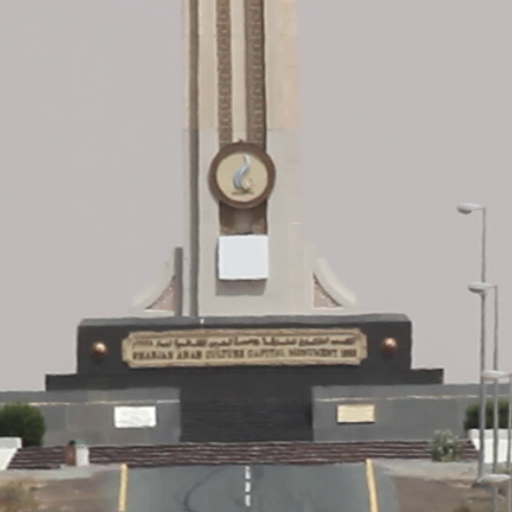}&Monument~\cite{CLEAR:data}&100&512$\times$512&Distant Monument, Outdoor, Static, Real, No-GT, Medium Turbulence\\
    \includegraphics[width=0.07\textwidth]{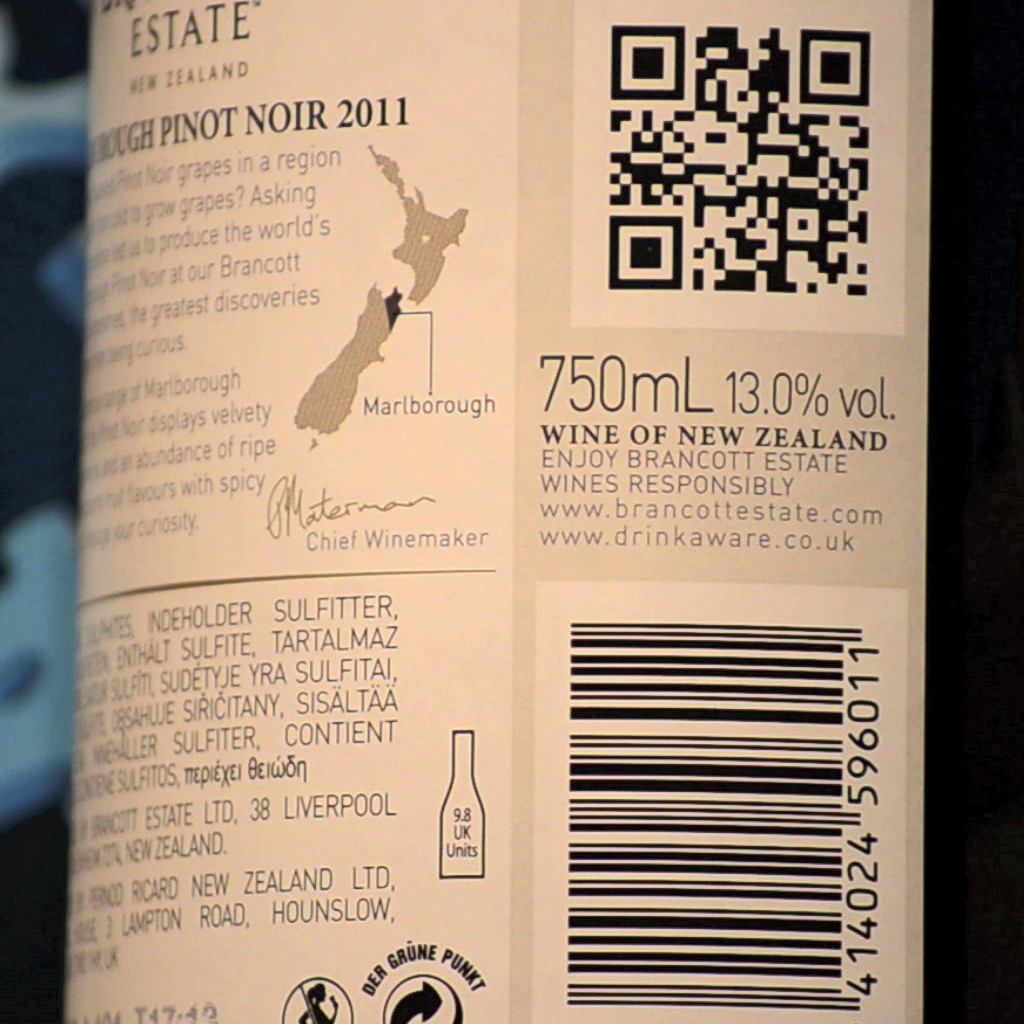}&Barcode~\cite{CLEAR:data}&100&1024$\times$1024&Single barcode (with text), Indoor, Static, Synthetic, GT, 3 Turbulence levels\\ 
    \includegraphics[width=0.07\textwidth]{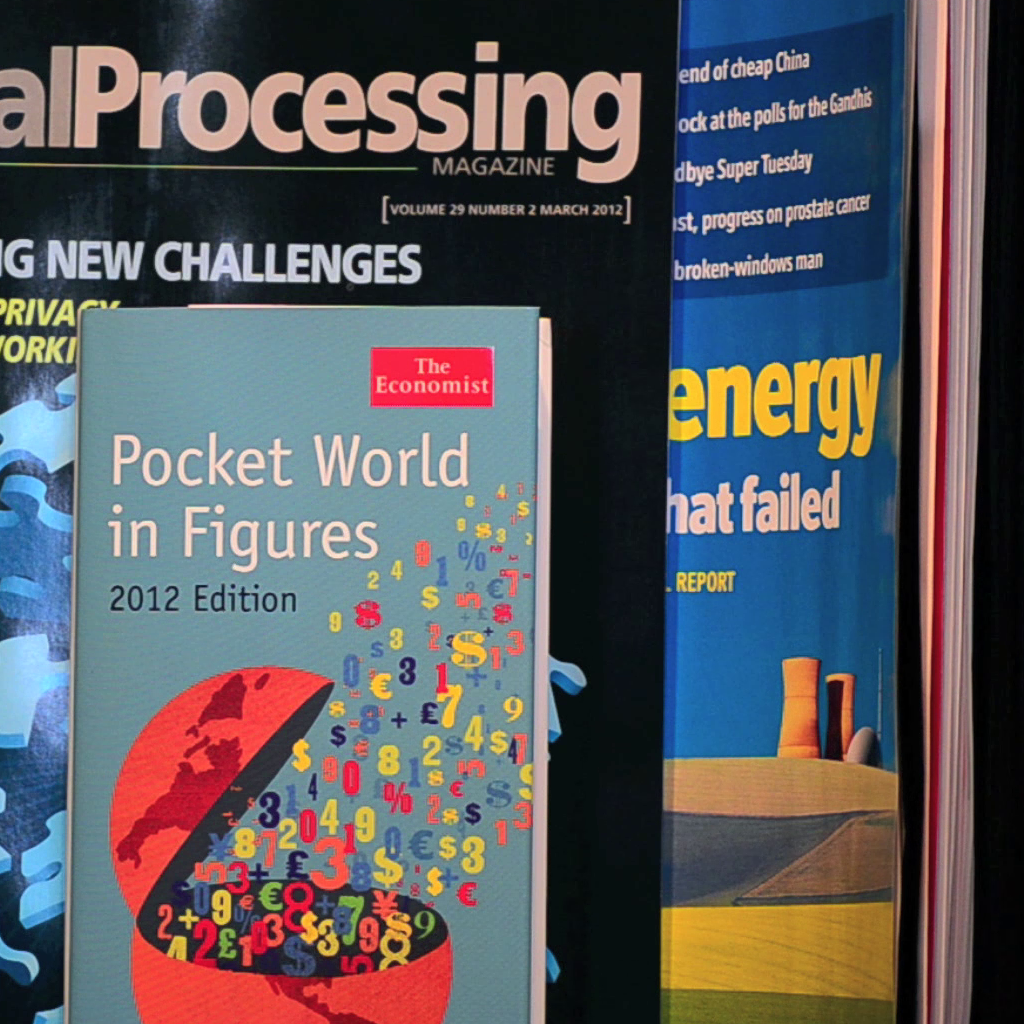}&Books~\cite{CLEAR:data}&99&1024$\times$1024&Three books, Indoor, Static, Synthetic, GT, 3 Turbulence levels\\ 
    \includegraphics[width=0.07\textwidth]{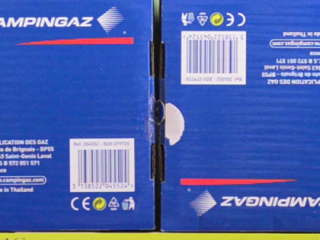}&Boxes~\cite{CLEAR:data}&100&320$\times$240&Two boxes, Indoor, Static, Synthetic, GT, 3 Turbulence levels\\ 
    \includegraphics[width=0.07\textwidth]{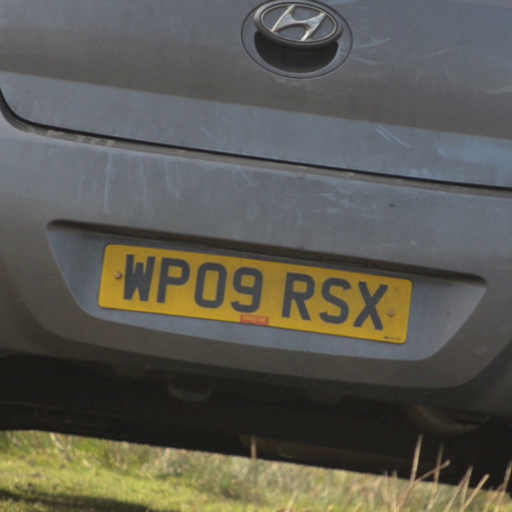}&CarBack~\cite{CLEAR:data}&100&256$\times$256&Single car, Outdoor, Static, Synthetic, GT, 3 Turbulence levels\\ 
    \includegraphics[width=0.07\textwidth]{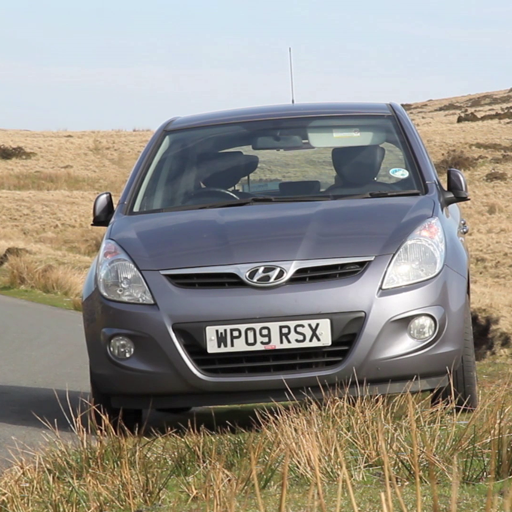}&CarFront~\cite{CLEAR:data}&99&512$\times$512&Single car, Outdoor, Static,  Synthetic, GT, 3 Turbulence levels\\ 
    \includegraphics[width=0.07\textwidth]{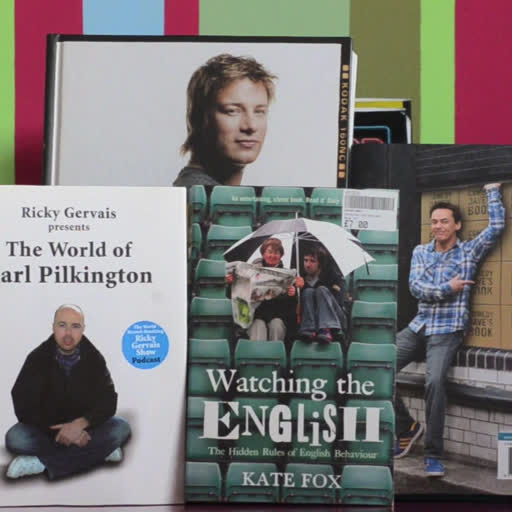}&Faces~\cite{CLEAR:data}&96&512$\times$512&Four books featuring faces, Indoor, Static, Synthetic, GT, 3 Turbulence levels\\ 
    \includegraphics[width=0.07\textwidth]{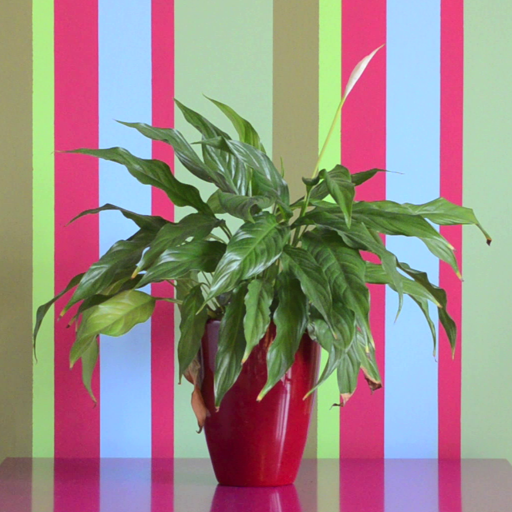}&Plant~\cite{CLEAR:data}&100&512$\times$512&Single plant, Indoor, Static, Synthetic, GT, 3 Turbulence levels\\ 
    \includegraphics[width=0.07\textwidth]{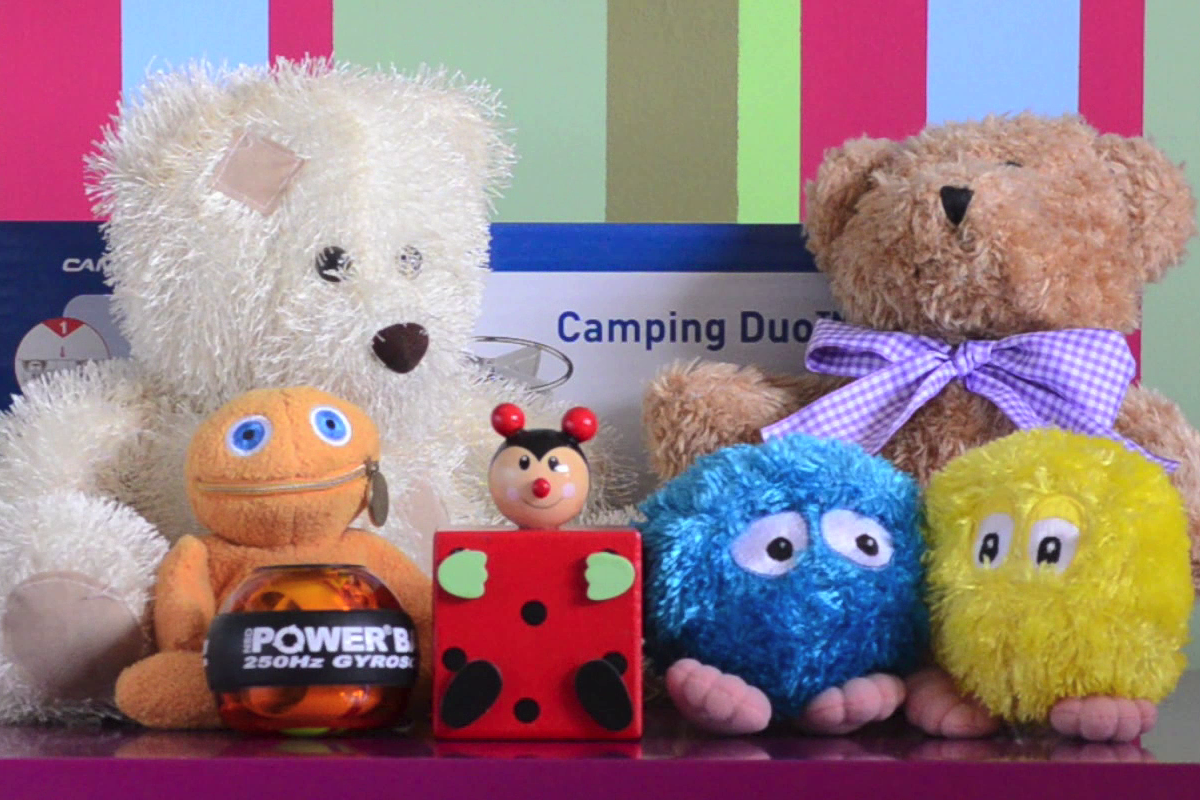}&Toys~\cite{CLEAR:data}&96&1200$\times$800&Six toys, Indoor, Static, Synthetic, GT, 3 Turbulence levels\\ 
    \includegraphics[width=0.07\textwidth]{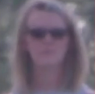}&BRIAR~\cite{10031155}&1,000&200$\times$50&Human, Outdoor, Static, Real, No-GT, Medium Turbulence\\ 
    \includegraphics[width=0.07\textwidth]{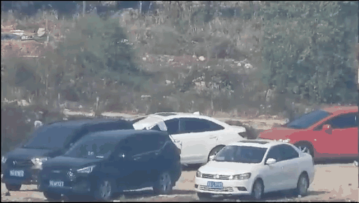}&RLR-AT~\cite{xu2024long}&800&1920$\times$1080&Cars, Buildings etc., Outdoor, Static camera, Real, No-GT, Medium to High Turbulence\\ 
    \hline
    \end{tabular}
    \end{table}
\clearpage
\subsection{Real datasets}
\label{ssec:realdata}

\subsubsection{Real datasets with ground truth}
To replicate turbulent conditions, Anantrasirichai[~\cite{Anantrasirichai:Atmospheric:2013,CLEAR:data}] utilised the heat generated by gas hobs to induce real turbulence in a controlled indoor environment. The different number of gas hobs used generates different levels of distortion levels. However, it is important to note that this method oversimplifies real atmospheric turbulence,  as the light rays propagate through the medium in very short distances compared to real outdoor atmospheric turbulence conditions. Additionally, fluctuations in humidity and the presence of small particles such as dust are not accounted for.  Furthermore, the dataset obtained from this method solely comprises static scenes.  \cite{Mao:accelaring:2021} also generated indoor real turbulence sequences using a number of ``heat chambers'' that are placed between the camera and a fixed scene (usually a printed card or poster).  The ground truth is generated by simply deactivating the heat chambers therefore not requiring post-processing to register the distorted and undistorted data.  The whole dataset is available in~\cite{Mao:data}.  

The Open Turbulent Image Set (OTIS) as presented in~[\cite{Gilles:Open:2017}] captures real atmospheric turbulence and provides both static and dynamic scenes (a subset of the OTIS dataset is shown in Table 1). The static scenes are captured from printed charts physically put in a hot outdoor environment at different distances from the camera to gain different levels of atmospheric turbulence.  The pseudo ground truth for the static scenes is generated through the registration of the digital image of the printed card with the average of the distorted video static sequence. The dynamic scenes do not contain any visual ground truth.  However, as the aim of the dynamic scene scenario is object tracking, the ground truth for these sequences is the frame-by-frame bounding boxes around the moving car toy.  Two separate outdoor real datasets with ground truth are the ``Chimney'' and ``Building'' sequences, generated by \cite{5540158}.  These two static sequences were captured on a platform through hot air exhausted by a building’s vent, which could be closed to generate ground truth images of the same scene.  
A recently released dataset~[\cite{xu2024long}] focuses on very long (1Km to 13Km) distance sequences with significant turbulence.  This dataset contains a large number (1500 sequences each with an average of 800 frames) of high resolution (1920$\times$1080) videos.

\subsubsection{Real datasets without ground truth}
Indoor scenes with moving toys and turbulence captured using an ``event camera'' were proposed by \cite{Boehrer:Turbulence:2021}. The turbulent air flows are created by a hot plate placed close to the camera.  Another example of a dynamic real dataset without ground truth is ``Car''[~\cite{Anantrasirichai:Atmospheric:2013,CLEAR:data}], in which a car travels towards the camera over a hot road that generates significant atmospheric turbulence.  Similarly, static real datasets without ground truth include ``Monument'' a static outdoor scene of a monument captured through atmospheric turbulence caused by heat.  Both ``Car'' and ``Monument'' are available on the University of Bristol's CLEAR website~[\cite{CLEAR:data}].  These two sequences were collected using a Canon EOS-1D Mark IV camera with a 400mm lens at an approximate distance of 500m at a temperature of 46°C in dry desert conditions. The atmospheric turbulence mitigation CVPR2022 challenge includes 50 sequences of hot-air images and 500 sequences of text patterns in its test datasets\footnote{http://cvpr2022.ug2challenge.org/track3.html}. The study by \cite{Jin:Neutralizing:2021} offers a real-world test set comprising 27 distinct dynamic scenes captured under relatively mild turbulence conditions, using hot roads as the medium.
\subsection{Synthetic Turbulence Generation}
\label{ssec:syndata}
\begin{figure}[h]
	\centering
 \includegraphics[width=0.5\columnwidth]{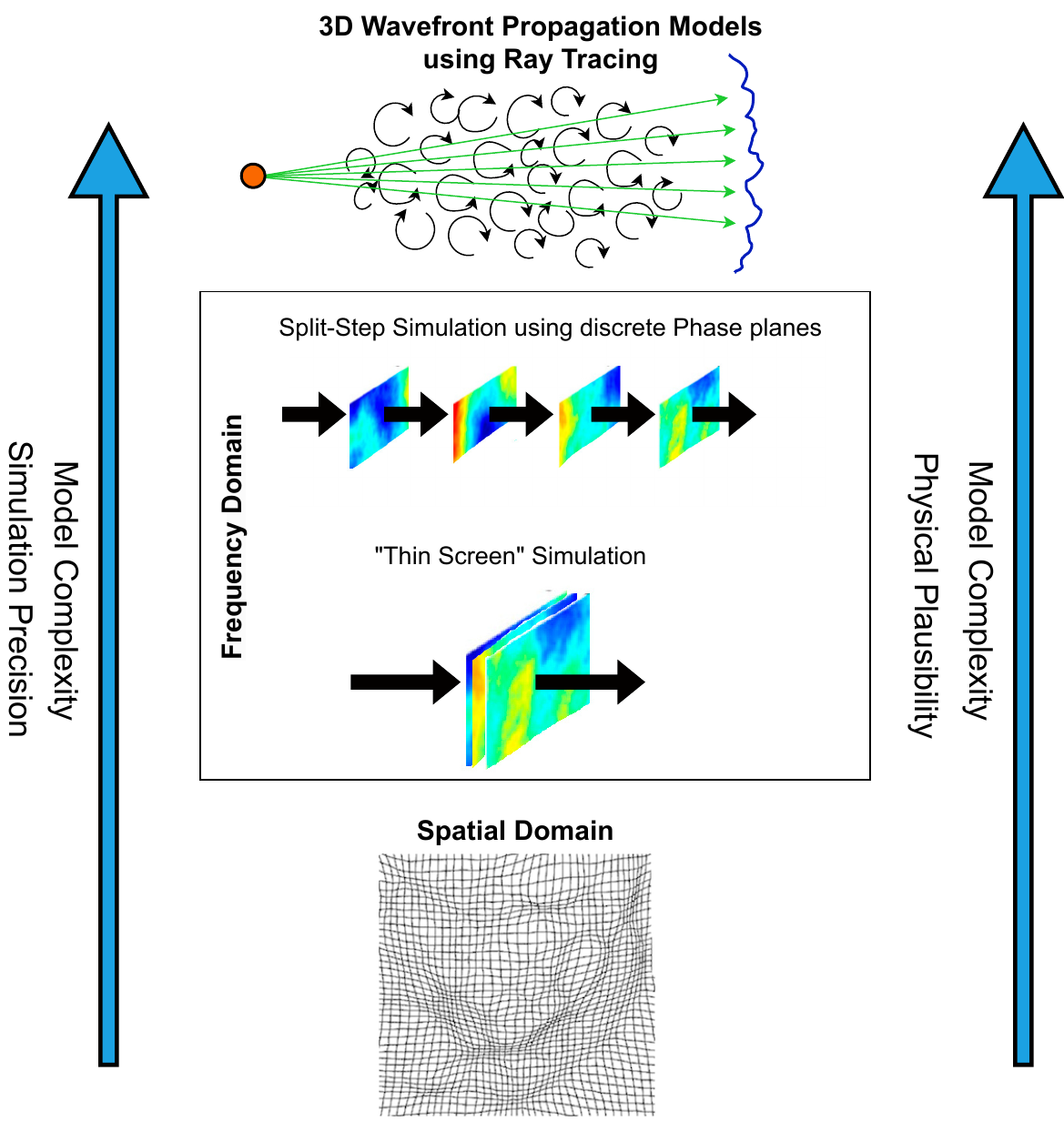}
					\caption{Non-Deep Learning based Synthetic Turbulence Simulation Methods}
    \label{fig:syndata1}
\end{figure}

Synthetic turbulence generation methods can be divided into four categories: 3D wavefront propagation models using ray tracing, spatial domain methods, frequency domain methods, and deep learning based methods.
The categorisation of the non deep learning methods is illustrated in Figure \ref{fig:syndata1}.  This figure shows the increasing levels of complexity (both of the model and required computational complexity) from na{\"\i}ve 2D processing to full 3D ray tracing solutions.  The simulation precision of the models also increases as more accurate and complex physical models are utilised.

\subsubsection{3D Wavefront Propagation Models using Ray Tracing}

The utilisation of physical models enables graphics engines to simulate the propagation of wavefronts through modelled atmospheric phenomena (e.g.\~[\cite{ihrke2007eikonal,gutierrez2006simulation}]). These models have proven to be effective in accurately representing atmospheric visual phenomena, including the Fata Morgana/Viking's End of the World and the Green Flash~[\cite{gutierrez2006simulation}]. Additionally, 3D rendering techniques have been developed to model complex lighting effects through refractive objects~[\cite{gutierrez2005non}]. However, these techniques are computationally complex and often require specialised hardware to generate videos within a practical amount of time. Currently, 3D ray tracing methods have not been used to generate artificial turbulence datasets for training data-based models.

\subsubsection{Spatial Domain}
The techniques described in~\cite{5540158} and \cite{Gao:Atmospheric:2019} use convolutions with a set of predefined basis functions, referred to as atmospheric blur kernels. These kernels are individually applied to overlapping patches, resulting in a spatially varying effect that covers the entire image. Formally this can be represented as:
\begin{equation}
    y = \sum_{p} \sum_k m_p w_k (h_k * x), 
    \label{eq:basis}
\end{equation}

\noindent where $x$ and $y$ are the clean and the distorted image, $h_k$ is the $k^\text{th}$ basis function, $w_k$ a spatially varying weight, (e.g.\ Gaussian weight), and $m_p$ represents a mask, where the pixel value equals to 1 if such pixel belongs to the $p^\text{th}$ patch, otherwise it equals to 0. This can be seen as a spatially varying weighted sum operation.
In~[\cite{Zhu:Removing:2013}], the distortion is caused by two PSFs: i) space-varying PSF (blur and turbulent motion), using a Gaussian function and B-spline model; and ii) space-invariant diffraction-limited PSF (sensor optics), simply generated with a disc function. Later, the authors in~[\cite{Hunt:synthesis:2018}] applied dictionary learning~[\cite{Aharon:KSVD:2006}] to a small set of real data. With this sparse representation, the PSFs (custom dictionary or basis functions) are estimated directly in the focal image plane. Simple motion maps are randomly generated and the clean image is warped accordingly in~[\cite{Chak:Subsampled:2021}], followed by Gaussian blur. This method is oversimplified (relative to real-world turbulence), and micro aberrations due to mixing signals are apparent.  To evaluate their GAN based turbulence mitigation method (LTT-GAN)~[\cite{mei2023ltt}] utilised a 2D-based artificial turbulence generation method they define as ElasticAug which combines Gaussian blur (as per (\ref{eq:basis})) with the Elastic Transformation defined by~\cite{simard2003best}.

\noindent \textbf{2D Image Processing Based Simulations}\\
2D-based turbulence generation is the simplest turbulence simulation method to implement and does therefore not require extensive computational power. Often, parametric 2D models are used. These parametric models are based on the analysis
of atmospheric phenomena and their empirical distortions~[\cite{potvin2007parametric, repasi2011computer}]. Other 2D-based methods, (including~[\cite{5540158, Gao:Atmospheric:2019,fishbain2007real,Zhu:Removing:2013}]),
use simple Gaussian random functions to generate image
distortion fields. 
All of these methods can give visually pleasing results, but they do not use any physically plausible model or use a set of consistent and physics-based distortion correlations.

\subsubsection{Frequency Domain}
Amongst frequency domain techniques, the simplest model has been used in remote sensing, where turbulent fluctuations of temperature and humidity cause atmospheric propagation delays of radar signals when travelling through the troposphere.  This turbulence is spatially correlated and the covariance in the phase domain can be described using an exponentially decaying function~[\cite{Anantrasirichai:deep:2019}]. The one-dimensional covariance function is defined as:

\begin{equation}
    c_{ij} = \sigma^2_{max} e^{-\kappa d_{ij}},
\end{equation} 
\noindent where $c_{ij}$ is the covariance between pixels $i$ and $j$, $d_{ij}$ is the distance between the pixels, $\sigma^2_{max}$ is the maximum covariance and $\kappa$ is the decay constant, which is equivalent to the inverse of the $e$-folding wavelength. In satellite imaging, this can be assumed to be radially symmetric and have a homogeneous structure across the image. However, in long-range imaging close to the ground,  turbulent mixing of the air occurs, resulting in three-dimensional spatial variation of the refractivity~[\cite{Grabner:Atmospheric:2011}].

A turbulent medium causes phase fluctuations, while the phases in the frequency domain of the image contain information of spatial displacement~[\cite{Chen:registation:2011}]. Many simulated methods therefore model atmospheric turbulence in the phase domain. For example, the authors in \cite{Woods:Lucky:2009} describe atmospheric turbulence effects with wavefront sensing principles, and employ phase diversity to estimate the level of turbulence severity.
The technique uses two phase-diverse images and measures the through-focus symmetry of the  PSF\@. The regions of the scene containing a small amount of atmospheric degradation will have the same PSF, whilst the regions which are imaged with significant aberration will have different PSFs in the two phase-diverse images.

Atmospheric turbulence was also modelled using the Power Spectral Density (PSD) of the refractive index first developed by \cite{Kolmogorov:local:1991}.  A commonly utilised model is a modified Von Karman spectra model~[\cite{Schmidt:Numerical:2010,Chatterjee:slit:2014,Hardie:simulation:2017}].
The anisoplanatic Point Spread Function (PSF) proposed by~\cite{Hardie:simulation:2017} is synthesised by means of wave propagation analysis. A sequence of phase screens is generated using the split-step method based on Fresnel diffraction, as illustrated in Figure \ref{fig:syndata1}. In the work by~\cite{Schwartzman:Turbulence:2017}, the Angle of Arrival (AoA) is randomly generated to introduce fluctuating motion. The distortion field, which is derived from the variance of the AoA with respect to focal length, distance, and $C_n^2$ parameters in the Fourier domain, appears to be realistic subjectively but does not account for any blurring effect of atmospheric turbulence.

Recently~\cite{Chimitt:Simulating:2020} have developed an optimised split-step method that combines the phase screens into a single and collapsed ``thin screen'' (see Figure \ref{fig:syndata1}.)  In their study, they modelled phase distortion caused by atmospheric turbulence as a function of frequency, focal length, aperture diameter and a random vector while utilising Zernike polynomials. Their findings indicate that phase distortion introduces a random tilt to the Point Spread Function (PSF), and the degree of tilt is associated with the random vector. The simulated images for subtle turbulence closely resemble the actual images, albeit subjectively. However, for strong turbulence, the simulated images exhibit higher spatial correlation and greater blur than the real images (see Figure 19 in~\cite{Chimitt:Simulating:2020}).  This method of synthetic turbulence generation (and its optimised version by~\cite{Mao:accelaring:2021}) has been utilised by~\cite{Cheng:Restoration:2023} by using a CNN-based model to develop a deep learning based mitigation method using ImageNet~[\cite{deng2009imagenet}] as the ground truth (and the synthetically distorted version of ImageNet as the network inputs).  The methods developed by~\cite{Chimitt:Simulating:2020} and~\cite{Mao:accelaring:2021} have been more recently further optimised to develop a real-time artificial turbulence generation system~[\cite{chimitt2022real}]. 

\subsubsection{Deep Learning Based Methods}

Recently, deep learning techniques have gained significant interest in generating artificial turbulence.   Generative models, such as Generative Adversarial Networks (GANs~[\cite{goodfellow2014generative}]), offer rapid processing and yield promising results. However, the primary limitation lies in the requirement of ground truth for the training process. For instance, the approach proposed by \cite{Rai:learning:2020} employs a simulated dataset introduced by \cite{Schwartzman:Turbulence:2017}.  Furthermore, GANs have also been utilised for artificial turbulence generation by~\cite{miller2021machine}. Deep learning techniques also circumvent certain complex steps in turbulent generation. \cite{Mao:accelaring:2021} utilise a few convolutional layers to transform Zernike polynomials~[\cite{Noll:Zernike:1976}] for high-order aberrations into the spatial domain (optimising the work developed by~\cite{Chimitt:Simulating:2020} and described below). They train the network using synthetic Point Spread Functions (PSFs) ranging from weak to strong turbulence levels.  

\subsubsection{Evaluation of Simulated Turbulence Methods}
Qualitatively, none of the simulated turbulence methods described above match the effects of turbulence of real sequences. 
Quantitatively, a comparison of a small number of simulated turbulence methods has been reported by \cite{nair2022comparison}.  This work trains two network based turbulence mitigation methods (AT-Net~[\cite{yasarla2021learning}], MPRNet~[\cite{zamir2021multi}]) using data generated using five different open source turbulence generation methods (i.e.\~[\cite{Chak:Subsampled:2021,Mao:accelaring:2021,Chimitt:Simulating:2020,Schwartzman:Turbulence:2017,mei2023ltt}]).  These five methods were used to add artificial turbulence to the FFHQ dataset~[\cite{karras2019style}].  Once trained, AT-Net and MPRNet were used to remove turbulence from the 
LRFID dataset~[\cite{miller2019data}].
The LRFID face dataset consists of real-world
turbulence distorted images of one hundred individuals in six different poses.  The ArcFace face recognition system~[\cite{deng2019arcface}] is used to quantitatively evaluate the turbulence-mitigated LRFID dataset using the combined ten methods described above (two mitigation methods combined with five turbulence simulation methods).  The conclusion from this paper is that the method developed by~[\cite{Chimitt:Simulating:2020}] provides the best quantitative and qualitative results in most cases (of the five artificial turbulence generation methods considered).

\subsection{Metrics for Performance Characterisation}
\label{ssec:metrics}
For any datasets with ground truth (or pseudo ground truth) distortion and perceptually inspired metrics are often utilised (e.g.\ PSNR, MS-SSIM, SSIM~[\cite{Bovik_SSIM}] and VMAF~[\cite{ramsook2021differentiable}]). As the majority of turbulence datasets do not have any ground truth, No Reference (NR) metrics are often used.  \cite{Anantrasirichai:Atmospheric:2013} identified a subset of NR metrics applicable to assessing the quality of turbulence mitigation, including JP2K~[\cite{sheikh2005no}], AQI~[\cite{Gabarda:blind:2007}] and QSVR~[\cite{Anantrasirichai:Atmospheric:2013}].  It has been recently argued that No Reference metrics for image and video quality lack accuracy and reproducibility~[\cite{pinson2022no}].

\section{Model-based Methods for Atmospheric Turbulence Mitigation}
\label{sec:turbModels}

A common mathematical model of atmospheric turbulence effects used in the literature is: 
\begin{equation}
y = \mathcal{D}x + n,   
\label{eq:turb1}
\end{equation}
\noindent where $x$ and $y$ are the ideal and observed images respectively, $\mathcal{D}$ represents unknown geometric distortion and blur, while $n$ represents noise. Despite being simple, the perfect solution is however in practice impossible since the problem is ill-posed.  Some authors (e.g.~[\cite{mao2022single,chimitt2022real,Mao:accelaring:2021}]) have further decomposed the transform $\mathcal{D}$ into separate geometric distortion and blur operations.  However useful this is conceptually, it is impossible to precisely separate these individual transforms and their closed-form separation ignores other combined distortion effects (caused by effects such as pollution, fog, and haze). 

Conventional methods have attempted to solve (\ref{eq:turb1}) by modelling $\mathcal{D}$ as a PSF
and then employing blind deconvolution with an iterative process to
estimate $x$~[\cite{Lam2000, Harmeling:online:2009}]. The results of assuming a space-invariant PSF usually exhibit artefacts. A patch-based technique was introduced by Hirsch to limit the space-invariant PSF within a small patch in~[\cite{5540158}].  This method reduces the space-invariant PSF artefacts significantly.
Furthermore, prior assumptions have been included to improve the estimation, e.g.\ the PSF is assumed to follow the Fried kernel~[\cite{Deledalle:blind:2020}].

\begin{figure}[t]
	\centering
      		 \includegraphics[width=0.65\columnwidth]{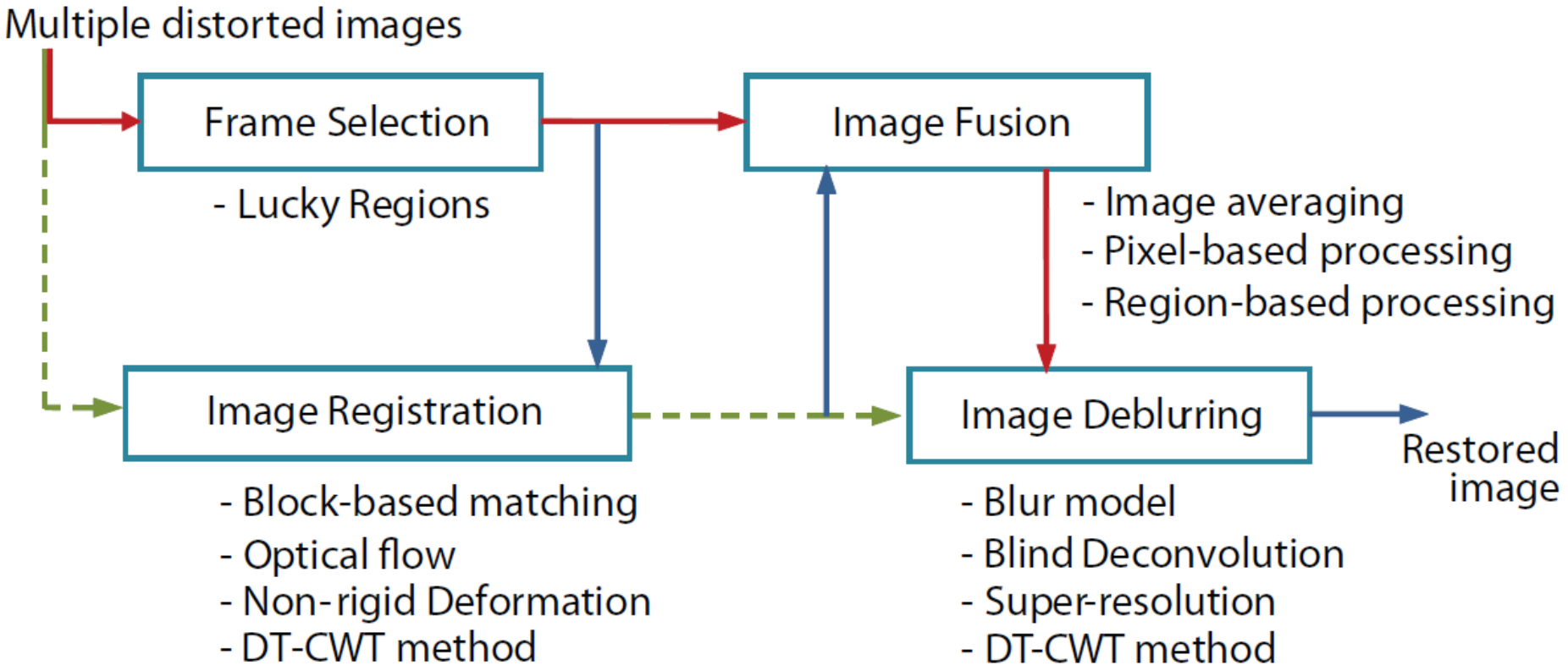}
					\caption{Common workflow of atmospheric turbulence removal using multiple frames~[\cite{Anantrasirichai:Atmospheric:2013}].}
    \label{fig:methoddiagram3a}
\end{figure} 

Removing the visible ripples and wave effects with a single image is almost impossible using model-based methods. Multiple distorted frames are employed to capture statistics of micro-spatial displacements among frames. A common workflow is shown in Figure \ref{fig:methoddiagram3a}, described in \cite{Anantrasirichai:Atmospheric:2013}. This workflow starts with frame selection or frame registration. As the atmospheric distortion always causes blur, the frame selection employs a sharpness metric to determine the amount of detail the image contains.
The registration process involves creating a reference frame so that the neighbouring frames are warped to it~[\cite{Anantrasirichai:Atmospheric:2013, Patel:adaptive:2019, Mao:Image:2020}].  However, this also creates a problem as a reference frame is unknown and has to be created from the distorted sequence. The simplest way to create the reference frame is temporal averaging of a number of frames in the sequence with an assumption that variation of micro displacement amongst frames is random. The registration process, (also called stabilisation in some papers~[\cite{Oreifej:Simultaneous:2013,Xue:Video:2016}]) attempts to align objects temporally to solve for small movements of the camera and temporal variations due to atmospheric refraction. The image fusion block may subsequently be employed in order to combine several aligned images~[\cite{Anantrasirichai:Mitigating:2012, Mao:Image:2020}]. Then a deblurring~[\cite{Zhu:Removing:2013, Mao:Image:2020}] or super-resolution process~[\cite{Nieuwenhuizen:Dynamic:2019}] is applied to the combined image (due to the blurring effect of the fusion stage).

When the videos contain moving objects in the turbulent atmospheric medium, most methods reconstruct the static background first and employ thresholding techniques using motions and/or intensity. 
\cite{Li:Suppressing:2009} and \cite{Delport:Scintillation:2010} detect moving objects by thresholding the motion error present after restoring the static background to distinguish the real motion and turbulent motion. \cite{Haik:Effects:2007} model the background by tracking local displacement across a number of successive frames.  Conversely,~\cite{Deshmukh:moving:2013} model the background using a Gaussian Mixture Model (GMM). \cite{Halder:geometric:2015} proposed an iterative approach to remove turbulent motion and the moving object is masked using simple thresholds on both motion and intensity. Unfortunately, this method does not remove the distortion around a moving object.  A low-rank matrix approach decomposing the distorted image into background, turbulence, and moving object method is proposed by~\cite{Oreifej:Simultaneous:2013}. An adaptive threshold technique applied to the background model was introduced by~\cite{Chen:detecting:2014}. The initial background is created from a temporal median filter, with the background updated every frame using a weighted average between the current frame and the background model. 

Most methods detect long-distance target objects that are at sufficiently low fidelity to exhibit little or no detail, instead appearing as blurred silhouettes or blobs ~[\cite{Haik:Effects:2007,Deshmukh:moving:2013, Oreifej:Simultaneous:2013, Chen:detecting:2014, Gilles:Detection:2018}].  A small number of methods in the literature were introduced to mitigate turbulent distortion on large moving objects. The moving objects are detected using block matching techniques~[\cite{Huebner:Software:2011}], smoothed motion trajectories~[\cite{Foi:methods:2015}], and a gradient pyramid [\cite{Anantrasirichai:Atmospheric:2018}]. The compensated moving areas are aligned in the 3D spatio-temporal volume and the turbulent distortion within these areas is suppressed in the same way as the static background areas.  \cite{Foi:methods:2015} estimate the true motion by smoothing the motion trajectories to remove small random movements caused by turbulence across a fixed number of successive frames. \cite{Kelmelis:practical:2017} developed the ``\textit{Dynamic Local Averaging}'' method, which determines the number of frames to employ for averaging, to avoid any unwanted effects. This method however may not mitigate the distortion on the moving objects as mostly only one frame is employed. Temporal filtering with local weights determined from optical-flow is employed to address this by \cite{Anantrasirichai:Atmospheric:2018}. However, artefacts in the transition areas between foreground and background regions can remain. A spatio-temporal non-local averaging technique was used by~\cite{Mao:Image:2020} to create a reference frame. If a patch of the moving object shows low similarity to its spatial-temporal neighbours then a lucky region fusion method is applied, followed by blind deconvolution.
Recently,~\cite{Boehrer:Turbulence:2021} investigated the use of an ``event camera'' (also termed a ``neuromorphic camera'') for turbulence mitigation.  Event cameras offer ultra-high framerate as they capture only logarithmic changes in local brightness. They tested the camera capturing sequences through turbulent air flows created by a hot plate. They applied optical-flow to segment a moving object and reduced the turbulence effects of the static background with recursive temporal averaging.

Within a matrix decomposition framework, various authors have utilised a low-rank based based prior of a static background to remove distortions~[\cite{wright2009robust,Oreifej:Simultaneous:2013,xu2024long}].  Most recently have combined such matrix based optimisation with a frequency-aware reference frame generation for better registration.  

In conclusion, the model-based approaches achieve desirable restoration performances of the static scenes. However, there are two major problems with this class of methods: i) high computational complexity, making it difficult for real-time applications; ii) artefacts from moving objects due to imperfect alignment created when combining multiple images.

\section{Deep Learning-based Methods for Atmospheric Turbulence Mitigation}
\label{sec:deep}

Deep learning is a subset of machine learning that employs Deep artificial Neural Networks (DNNs). The word ``deep'' in this context means that there are a large number of hidden layers of neuron collections that have learnable weights and biases. When the data being processed occupies multiple dimensions
(images for example), Convolutional Neural Networks (CNNs) are often employed~[\cite{Anantrasirichai:Artificial:2021, Goodfellow-et-al-2016}]. CNNs
are (loosely) a biologically-inspired architecture and their results are tiled so that they overlap to obtain a better representation of the original inputs. Recurrent Neural Networks (RNNs) have been widely employed on sequential data~[\cite{Goodfellow-et-al-2016}]. CNNs extract spatial features from input images using convolutional filters whereas RNNs extract sequential features in time-series data using memory cells. 

This section further divides into two subsections to discuss the existing approaches specifically proposed to mitigate atmospheric turbulence (Section \ref{ssec:existingmethods}) and to examine the performance of existing approaches proposed for other applications when transferred to atmospheric turbulence problems (Section \ref{ssec:transfer}).

\subsection{Methods proposed for atmospheric turbulence removal}
\label{ssec:existingmethods}

Similar to many other applications, deep learning technologies have gained attention in atmospheric turbulence mitigation, but are still in their early stages of adoption. The earliest work in the literature~[\cite{Gao:Atmospheric:2019}] employed the widely used standard CNN-based Gaussian denoising method: DnCNN~[\cite{Zhang:dncnn:2017}]. The method restores a sequence on a frame-by-frame basis but also concludes that the average of multiple frames gives a better result than using a single frame. In \cite{Mao:accelaring:2021}, a simple, popular UNet architecture~[\cite{Ronneberger:Unet:2015}] (originally proposed for medical image segmentation) has been trained with 5000 simulated sequences, achieving promising results. However, the input of the network is a 50-frame concatenated volume to give a single restored output; hence, this approach is not feasible for distorted videos with moving objects.  Some proposed deep learning based turbulence mitigation methods follow a traditional workflow, i.e.\ first the reference frame is constructed, a series of frames are registered to the reference frame, they are averaged and deep learning-based deblurring is applied, e.g.\ using DnCNN~[\cite{Nieuwenhuizen:Deep:2021}].  More complex encoder-decoder structures have also been investigated by~\cite{Cheng:Restoration:2023} and~\cite{anantrasirichai2023atmospheric}. AT-Net~[\cite{yasarla2021learning}] employed two UNet architectures, where the first one estimates a distortion map from multiple frames, and the second one uses this distortion map to remove geometric blurs.

As an exact ground truth is unavailable for the atmospheric turbulence problem, a self-supervised technique has been proposed by \cite{Li:Unsupervised:2021}, where geometric distortion is removed using a grid-based rendering network. The method estimates spatial displacements and reconstructs the distorted frames. The clean frame is then the output of when the zero-displacement map is fed. This method however requires multiple distorted frames, so it works only for static scenes.

\subsubsection{GAN Based Systems}
A WGAN (Wasserstein Generative Adversarial Network~[\cite{arjovsky2017wasserstein}]) was employed by \cite{Chak:Subsampled:2021}. The input consists of multiple lucky frames fed into a UNet generator. This appears only to work well for static scenes.  GANs have also been utilised to restore images of faces that have been corrupted by atmospheric turbulence~[\cite{lau2020atfacegan,mei2023ltt}.
\cite{Wang:deep:2021}] proposed a framework comprised of two CNNs for aberration phase reconstruction and Zernike coefficients reconstruction. We cannot comment on the performance of this work as only one small image result is given by the authors.
 The TSR-WGAN introduced by \cite{Jin:Neutralizing:2021} incorporates both temporal and spatial information within the three-dimensional input to learn the representation of the residual between the observed and latent ideal data. Recently, an RNN has been integrated into the generator of GAN to predict optical flow field for moving objects~[\cite{s23218815}]. The results are sharper than those achieved by AT-Net~[\cite{yasarla2021learning}]. Recently~\cite{wang2024real} utilised a GAN based model to bridge the gap between real and synthetic atmospheric distortion using a Teacher-Student method. 
 Further details on using GANs for atmospheric turbulence removal can be found in~[\cite{Cheng:research:2023}].

\subsubsection{Transformer / Attention Based Systems}
Most of the previously discussed work either uses CNNs, GAN or encoder-decoder (e.g.\ UNet) architectures.  \cite{mao2022single} have shown that the static filter weights and limited filter support of CNNs are not able to effectively model the spatial dynamics caused by atmospheric turbulence.  Within this work, they utilise an encoder-decoder backbone structure similar to that of UNet but use transformer layers instead of convolutional layers.  Furthermore, they use a physics-inspired head (comprised of a reconstruction block and a degradation block) that inputs the features output by the transformer backbone and gives the reconstructed output.  This network is termed TurbNet by~\cite{mao2022single}.

\cite{zhang2024imaging} proposed a Turbulence Mitigation Transformer (TMT). The framework utilised a CNN UNet architecture that inputs a small window of contiguous frames in order to remove spatial distortions, the output of which is input into a transformer-based network to remove blur. Recently, they proposed a new method ~[\cite{Zhang:Spatio:2024}] that replicates the transitional workflow, comprising feature-to-reference registration, temporal fusion, and post-processing. The concept of an attention module used in transformers was applied to feature registration via the deformable attention alignment block.  Furthermore a multi-head temporal-channel self-attention method was used for temporal fusion.  \cite{wu2024semi} have also recently proposed a semi supervised / self attention based system that utilises a ``mean teacher'' [\cite{tarvainen2017mean}] method of utilising unlabelled data.  They are able to demonstrate that basis functions created within the trained network exhibit spatial correlations, visually similar to the Zernicke polynomials utilised by work such as~[\cite{Chimitt:Simulating:2020}] and~[\cite{Noll:Zernike:1976}].  \cite{jaiswal2023physics} have utilised an attention based network (Swin) combined with a diffusion for reportedly state of the art results. Finally, a transformer based network was utilised by \cite{Xia_2024_CVPR} that used a single colour frame together with an accompanying narrow band image.

\textbf{3D Spatio-Temporal Processing using Transformers}
The majority of the previously discussed methods have used 2D processing.  Transformer based UNets for the segmentation of medical volumetric sources have included the 3D versions of the Mamba~[\cite{gong2024nnmamba}] and SWIN~[\cite{yang2023swin3d}] transformer variants.

These methods have been adapted to be used with spatio-temporal 3D blocks within the experiments detailed below.

\subsubsection{Methods Based  on Denoising Diffusion Probabilistic Models}
Denoising diffusion probabilistic models have very recently been used to mitigate turbulence-distorted faces by \cite{nair2023ddpm}, called AT-DDPM.  It is anticipated that parallel diffusion models developed for blind inverse problems~[\cite{chung2023parallel}] and denoising-based diffusion (ILVR) methods~[\cite{choi2021ilvr}] would be applicable to turbulence mitigation.
AT-VarDiff~[\cite{Wang:Atmospheric:2023}] includes task-specific prior information, extracted by a variational autoencoder, into the conditional diffusion model. This produces sharper results than AT-DDPM.


\begin{table*}[!ht]
	\centering
	\caption{Existing turbulence mitigation methods and their properties}
	\small
\hspace*{-2.7cm}
		\begin{tabular}{cccc}
		\hline
		 \textbf{Methods} & \textbf{static scene} & \textbf{small object} & \textbf{large object} \\
			\hline
            & & & \\
		\multirowcell{7}{model-based} & \multirowcell{7}{\cite{Anantrasirichai:Atmospheric:2013};\\ \cite {Zhu:Removing:2013};\\ \cite{Xue:Video:2016}; \\ \cite{Patel:adaptive:2019}; \\ \cite{ Deledalle:blind:2020}}  & \multirowcell{7}{\cite{Oreifej:Simultaneous:2013}; \\ \cite{Chen:detecting:2014}; \\ \cite{ Halder:geometric:2015}; \\  \cite{Zhang:Stabilization;2018}; \\ \cite{Elahi:Detecting:2018}; \\ \cite{xu2024long}} & \multirowcell{7}{\cite{Foi:methods:2015}; \\ \cite{Anantrasirichai:Atmospheric:2018}; \\ \cite{ Nieuwenhuizen:Dynamic:2019}; \\ \cite{Mao:Image:2020}; \\ \cite{Boehrer:Turbulence:2021}} \\
            & & & \\
            & & & \\
            & & & \\
            & & & \\
            & & & \\
            & & & \\
            \hline
		\multirowcell{15}{learning-based} & \multirowcell{15}{\cite{mao2022single}; \\ \cite{Vint:analysis:2020}; \\ \cite{Chak:Subsampled:2021}; \\ \cite{nair2023ddpm}; \\ \cite{Mao:accelaring:2021}; \\ \cite{Jin:Neutralizing:2021}; \\ \cite{Cheng:Restoration:2023}; \\ \cite{Wang:deep:2021}; \\ \cite{lau2020atfacegan}; \\ \cite{yasarla2021learning}; \\ \cite{Li:Unsupervised:2021}$^*$; \\ \cite{Jiang_2023_CVPR}$^*$; \\ \cite{chung2023parallel}$^*$; \\ \cite{Wang:Atmospheric:2023}$^*$} &  & \multirowcell{15}{\cite{Nieuwenhuizen:Deep:2021}; \\ \cite{mei2023ltt}; \\ \cite{zhang2024imaging}; \\ \cite{Gao:Atmospheric:2019}; \\ \cite{anantrasirichai2023atmospheric}; \\ \cite{Zhang:Spatio:2024}; \\ \cite{Xia_2024_CVPR}; \\ \cite{wu2024semi}; \\ \cite{wang2024real}; \\ \cite{jaiswal2023physics}} \\
            & & & \\
            & & & \\
            & & & \\
            & & & \\
            & & & \\
            & & & \\
            & & & \\
            & & & \\
            & & & \\
            & & & \\
            & & & \\
            & & & \\
            & & & \\
            & & & \\
            & & & \\
				\hline 
				$^*$ unsupervised learning
		\end{tabular}
	\label{tab:performance}
\end{table*}

\subsection{Relevant deep learning architectures}
\label{ssec:transfer}

Spatial displacement occurring due to atmospheric turbulence may be considered as noise. The residual derived from subtracting the distorted signal from the clean signal however does not have a simple Gaussian distribution, so simple Gaussian denoising methods such as DnCNN do not perform well for complex scenes~[\cite{Gao:Atmospheric:2019}]. Intensity fluctuations are better modelled with a Gaussian Mixture Model (GMM) following the tilt-variance statistics of atmospheric turbulence~[\cite{Hook:scene:21}]. A GMM may be integrated at the last layer of the neural networks~[\cite{Variani:GMM:2015}].
Geometric distortion is obviously the most challenging component of atmospheric turbulence to remove. However, as discussed in previous sections, multiple frame compensation of the micro displacements (the quasi-periodic property) results in a deblurring problem with spatial variation of unknown PSF. 

This section describes the possibility of using existing methods proposed to address inverse problems, e.g.\ image and video restoration, with both supervised learning and self-supervised learning approaches. 

\subsubsection{Supervised learning}
\label{sssec:Suplearning}

As supervised learning requires ground truth for training, the models described in this section are trained with synthetic datasets and real datasets with pseudo ground truth.
Here we target the distorted sequences with moving objects. \cite{Vint:analysis:2020} created a study of different deep learning model architectures used to remove turbulence within static scenes. In this study the authors tested several deep learning architectures with the synthetic examples created using the methods in~[\cite{Schmidt:Numerical:2010,Hardie:simulation:2017}], with three levels of distortions. They reported that Residual Dense Networks (RDNs)~[\cite{Zhang:RDN:2018}] gave the best performance in terms of PSNR. The RDN framework merges dense blocks, introduced in~[\cite{Huang:Densely:2017}], with residual blocks, used in ResNet~[\cite{He:ResNet:2016}].

A considerable number of deep learning models have been created to solve inverse imaging problems. We tested the following methods, which are state-of-the-art models that have either been used or considered to have potential in application to the problem of atmospheric turbulence mitigation.  

\begin{itemize}
    \item \textbf{FFDNet}~[\cite{zhang2018ffdnet}] offers the best denoising performance reported in many surveys~[\cite{Anantrasirichai:Artificial:2021,Tian:Deep:2020}].  FFDNet is an updated version of DnCNN~[\cite{Zhang:dncnn:2017}].  Both of these models use a simple collection of non-expanding or contracting convolutional layers that learn the noise when trained with a corrupted dataset (together with ground truth).   
    \item \textbf{UNet}~[\cite{Ronneberger:Unet:2015}] was originally created for medical segmentation applications. It features an encoder-decoder structure that has been used for many different image-to-image processing applications including turbulence mitigation (e.g.[~\cite{Gao:Atmospheric:2019,Mao:accelaring:2021,anantrasirichai2023atmospheric,zhang2024imaging}]). The basic encoder-decoder architecture of UNet has been extended to the use of complex layers~[\cite{anantrasirichai2023atmospheric}] and transformer layers~[\cite{mao2022single}]. 
    \item \textbf{Pix2Pix}~[\cite{isola2017image}] is an image-to-image translation architecture that uses a conditional GAN that requires a paired dataset to convert one type of image to another (i.e.\ cartoon to real image).
    \item \textbf{Deformable Convolutions} have been proposed to address geometric-temporal distortions (e.g. EDVR~[\cite{Wang:EDVR:2019}] and DLEFNet~[\cite{yang2023deformable}]).
    \item \textbf{Transformers} with a self attention mechanism~[\cite{vaswani2017attention}] was recently generalised for use with images in the ViT framework~[\cite{dosovitskiy2020image}].  Visually based transformers have already found application to turbulence mitigation (see above) and it is presumed that more recently developed methods such as SwinIR~[\cite{liu2021swin}] will also find effective application in this area.
\end{itemize}

\begin{figure}[!ht]
	\centering
      		 \includegraphics[width=0.75\columnwidth]{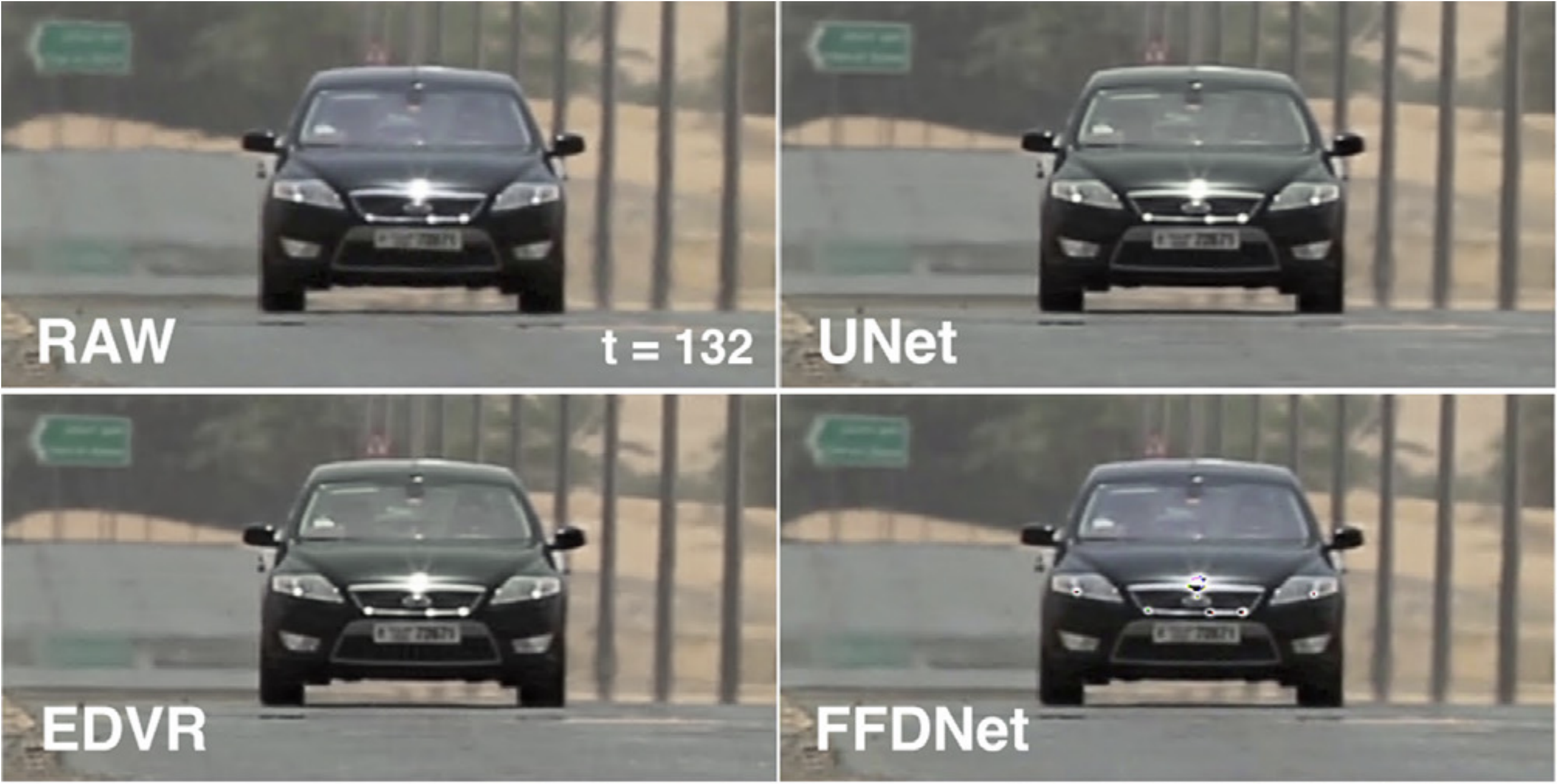}
					\caption{Result comparison of ``Car" sequence using UNet, EDVR and FFDNet~[\cite{anantrasirichai2023atmospheric}]}
    \label{fig:compare_inverse_problem}
\end{figure}

\noindent \textbf{Experimental setting}: Following~\cite{anantrasirichai2023atmospheric}, 12 sequences were used (each contained atmospheric turbulence and did not have any ground truth).  These sequences were \{Car, Van, Sony1, Sony2, Vay, Shutter, Dodge, Airport, NumberPlate, Secret, and Train\} as shown in Table 1 (Sony1, Sony2, Shutter, Dodge, Airport, Train, Secret, Men and Vay are omitted from the table due to licensing).
 The number of frames within each sequence was capped at 200 in order to prevent an unbalanced dataset. The amount of atmospheric turbulence varies from subtle to strong distortions as shown in Table 1.
   
 In order to model, process, and analyse the temporal effects of atmospheric turbulence, a sliding window method was adopted.  This sliding window takes a group of frames surrounding the current frame as input and outputs a single frame.  This is repeated across the entire sequence (in training and testing). As discussed by~\cite{anantrasirichai2023atmospheric}, five neighbouring frames were found to be the optimum number of frames, in terms of quantitative PSNR results.
  
Frame-by-frame pseudo ground truth was created for each of the 12 sequences using the model based CLEAR system~[\cite{anantrasirichai2023atmospheric,Anantrasirichai:Atmospheric:2013}]. A separate trained model was tested on each of the sequences in turn. Each separate model was trained on the entire dataset excluding the sequence to be tested.  This method was used due to the small size of the entire dataset and was also used within the system developed by~\cite{anantrasirichai2023atmospheric}.
The models were trained for 200 epochs with no early stopping.
During the training, spatial regions input were randomly cropped from the input sequences (of size $256\times 256$). This covered a larger and randomised spatial variation in the training dataset.
The Adam optimiser~[\cite{kingma2014adam}] was employed with an initial learning rate optimiser of 0.0001. All of the methods use default training and testing settings where applicable.

\noindent \textbf{Comparative Results}:
\label{sssec:Indicative Results}
Table \ref{tab:supresults} shows results for a range of turbulence mitigation techniques. The following shows the average results. 
For each of these test sequences models were trained on all of the remaining datasets (i.e.\ 11 training sequences for each of the testing sequences).  The first row indicates the PSNR and SSIM metrics between the raw input and the CLEAR output.
 The SwinIR, FFDNet and pix2pix methods process the sequences one frame at a time. The SwinIR, combining CNN architecture with transformer, performs the best among the three. The EDVR and UNet methods accept multiple-frame input, so they process the sequences using a sliding window as described above. These methods both utilise 5 levels of convolution (when using inputs of 256$\times$256 as described above).

 The TMT method uses the default settings of the TMT method~[\cite{zhang2024imaging}] utilising a sliding window that not only references but generates a range of output frames.  Such 3D data processes lends itself to 3D processing units.  The deblurring network within TMT is therefore replaced with 3D Mamba and SWIN based UNets to generate the methods TMT-Mamba and TMT-SWIN respectively. 
 The subjective results of some of these restoration methods are shown in Figure \ref{fig:compare_inverse_problem}.

\begin{table}[!ht]
\caption{Results for Supervised Methods for Turbulence Mitigation}
\label{tab:supresults}
\centering
\begin{tabular}{lcc}
\toprule
Method & PSNR & SSIM  \\
\hline
Raw distorted inputs & 26.21 &  0.8414 \\
UNet~\cite{Ronneberger:Unet:2015}     &28.49 & \underline{0.8786}  \\
Pix2Pix~\cite{isola2017image} & 25.52  & 0.7920 \\
FFDNet~\cite{zhang2018ffdnet}  & 26.93 & 0.8010 \\
EDVR~\cite{Wang:EDVR:2019}    & 28.85 & \textbf{0.8792}  \\
SwinIR~\cite{liu2021swin}  & 28.14  &  0.8649       \\
TMT~\cite{zhang2024imaging}  & 28.48 & 0.8652 \\
TMT~\cite{zhang2024imaging}+Swin3D~\cite{gong2024nnmamba} & \underline{29.09} & 0.8605\\
TMT~\cite{zhang2024imaging}+Mamba3D~\cite{yang2023swin3d}&  \textbf{29.21} & {0.8688}  \\
\bottomrule
\end{tabular}
\end{table}

\subsubsection{Unsupervised learning}
\label{sssec:Unsuplearning}

\begin{figure}
    \centering
    \includegraphics[width=0.7\textwidth]{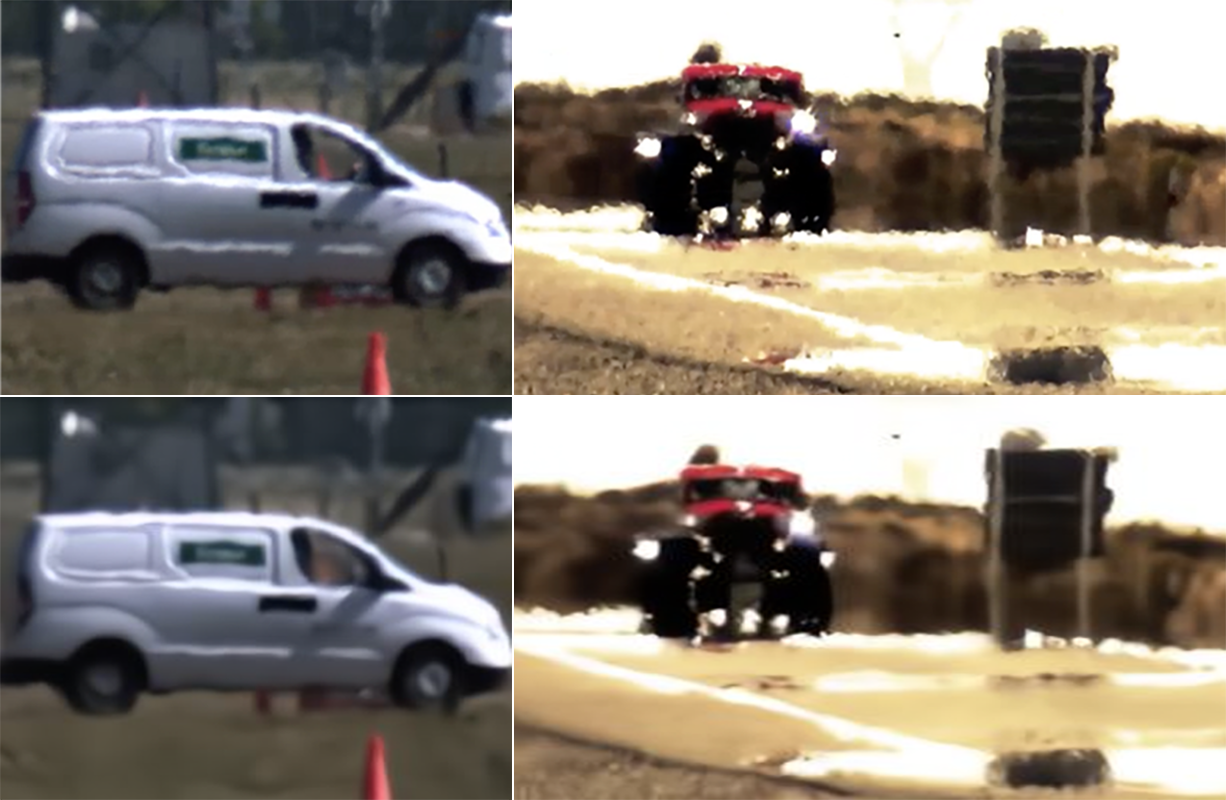}
    \caption[Caption for LOF]{Comparison of restored results (bottom row) using DDNM~[\cite{wang2022zero}] with the original distorted frames (top row)}
    \label{fig:results_DDNM}
\end{figure}

The main challenge of atmospheric turbulence restoration is the unavailability of ground truth in real scenarios. Although synthetic datasets and pseudo ground truth are employed in the training process, they are often over-simplified resulting in a lack of model robustness. Self-supervised learning techniques are therefore investigated here. Self-supervised learning models the problem through the loss function calculated from the data itself. This way, it can leverage the underlying structure in the data. Some potential methods are as follows:

\begin{itemize}
    \item \textbf{Deep Visual Prior}~[\cite{ulyanov2018deep, lei2020dvp,li2023random}] methods exploit neural networks, randomly initialised, to learn data prior. Three popular techniques are: i)  \textit{Deep Image Prior}~[\cite{ulyanov2018deep}] 
   is an unsupervised denoising method that uses a CNN to recompose an image from a latent and random input.  The authors of this work realised that convolutional architectures more easily learn and represent real image content compared to noise and distortions.  The deep image prior method therefore truncates the learning process after the underlying image content has been learnt but before the architecture is able to model the noise.
    ii) \textit{Deep Video Prior}~[\cite{lei2020dvp}] 
   is similar in method to the deep image prior method but uses videos instead of images (it still uses a CNN architecture).  
    iii) \textit{Deep Generative Prior}~[\cite{pan2021exploiting}] 
     effectively utilises an image prior within a  Generative Adversarial Network (GAN) structure for diverse applications in image restoration and manipulation.
    \item \textbf{Noise2Noise}~[\cite{lehtinen2018noise2noise}] 
   has the principle that given enough data, the clean targets can be replaced by other corrupted targets resulting in a learned model that is as effective as using clean targets.  It is conjectured that static atmospheric turbulence can be mitigated through the use of Noise2Noise methods (given the assumption that the scene is static and the turbulence can be considered as noise).  It is more difficult to ascertain how it could be used with turbulent dynamic scenes.
   The basic idea of Noise2Noise  was extended and generalised by Noise2Void~[\cite{krull2019noise2void}], Noise2Self~[\cite{batson2019noise2self}] and further refined in Self2Self~[\cite{quan2020self2self}].
    \item \textbf{Denoising Diffusion Probabilistic Model (DDPM)}~[\cite{wang2022zero}], originally proposed by \cite{Ho:Denoising:2020}, involves injecting noise into an image, training the model to predict the distribution of the next less noisy image at each step, and then using this knowledge to generate high-quality images by iteratively denoising pure noise inputs. It has proven effective for image generation tasks, reducing noise, and enhancing image quality in applications like image denoising and synthesis.
      Zero-Shot learning (ZSL) is utilised to perform unsupervised learning, where a model is trained to recognise data it has never seen during the training phase~[\cite{wang2022zero}]. The results of atmospheric turbulence mitigation are shown in Figure \ref{fig:results_DDNM}.
       The computational speed is very fast; however, the results are oversmoothed, and some ripple effects are still present.

    \item \textbf{Implicit Neural Representations} (INRs)
       are able to represent images using continuous functions within a simple Multi-Layer Perceptron (MLP) structure (e.g.~[\cite{sitzmann2020implicit}])
       NeRT (NEural Representations of Turbulence~[\cite{Jiang_2023_CVPR}]) adopts a ``tilt then blur'' structure of $\mathcal{D}$ in (\ref{eq:turb1}) (as analysed by \cite{chan2022tilt}). NeRT utilises both an INR representation together with an NDIR~[\cite{Li:Unsupervised:2021}] inspired deformed grid method to mitigate spatio-temporal tilts.  This paper claims state-of-the-art performance compared to other recently developed techniques. 
\end{itemize}

\footnotetext{Figures by Lingtao Cai, ``Low-light low-resolution image enhancement An end-to-end model'', BSc Computer Science, University of Bristol, 2023}

\section{Model Analysis}
\label{sec:modelA}
In order to characterise the function and utility of different parts of a model trained to mitigate atmospheric turbulence we provide an analysis of the energy of each model feature map for a range of simple models and a range of hyper-parameters. Here we performed analysis using a basic UNet-CNN architecture.

\subsection{Feature Map Energy}
When an input image (or range of images) is input into a trained model for testing (or inference) there is a huge number of intermediary features to potentially analyse within the network.  Rather than analysing all of the layer outputs within a model we analyse the direct outputs of the convolutional layers (Conv2D in PyTorch).  There are many options for measuring the activity of the output of the convolutional layers.  We choose the following energy measure for each of the convolutional stages (given that a Conv2D output is of dimension $C\times P\times Q$: Channels, Width and Height). The energy value serves as an indicator of the significance of specific features or patterns within the input data. Higher energy values indicate more substantial activations in response to particular features or patterns.

\begin{equation}
    E(i) = \frac{\sum_c\sum_x\sum_y |F_i(c,x,y)|}{C\cdot P\cdot Q}
\end{equation}

\noindent where $E(i)$ is the ``energy'' of the $i$th convolutional layer (i.e.\ Conv2D output), and $F_i(c,x,y)$ is the feature value output at spatial position $x,y$ in channel $c$ for layer $i$. 

Although energy most often refers to mean square than mean absolute values, we take the definition as the latter following the work by~\cite{park2018accelerating}.  We use mean absolute values as it gives less bias to large valued outliers in the analysis of feature-map outputs.

\subsection{Structure Analysis }
The figures shown below order the feature-map outputs from input to output (left to right) starting with larger (in spatial resolution) feature maps near the input and output with gradually decreasing sizes towards the point where the encoder and decoder structures join.
Furthermore, as the training and testing are done on a sliding window structure these figures show the ``energy'' values as distributions within the boxplots (the boxes show the quartiles of the distribution with the median in the centre). 

\begin{figure}
     \centering
      \begin{subfigure}[b]{0.31\textwidth}
         \centering
         \includegraphics[width=\textwidth]{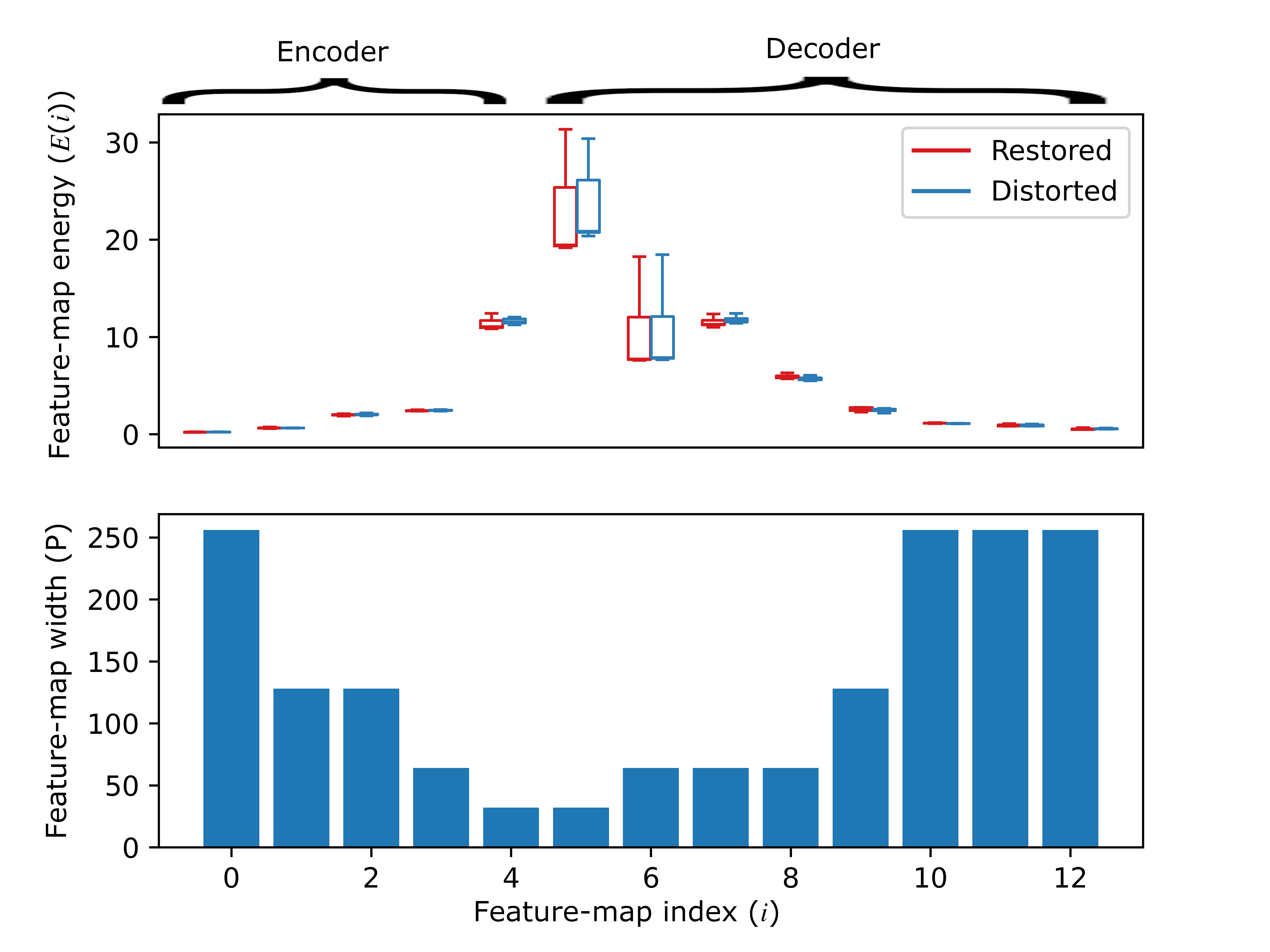}
         \caption{UNet-3 Levels}
         \label{fig:unet3}
     \end{subfigure}
     \hfill
     \begin{subfigure}[b]{0.31\textwidth}
         \centering
         \includegraphics[width=\textwidth]{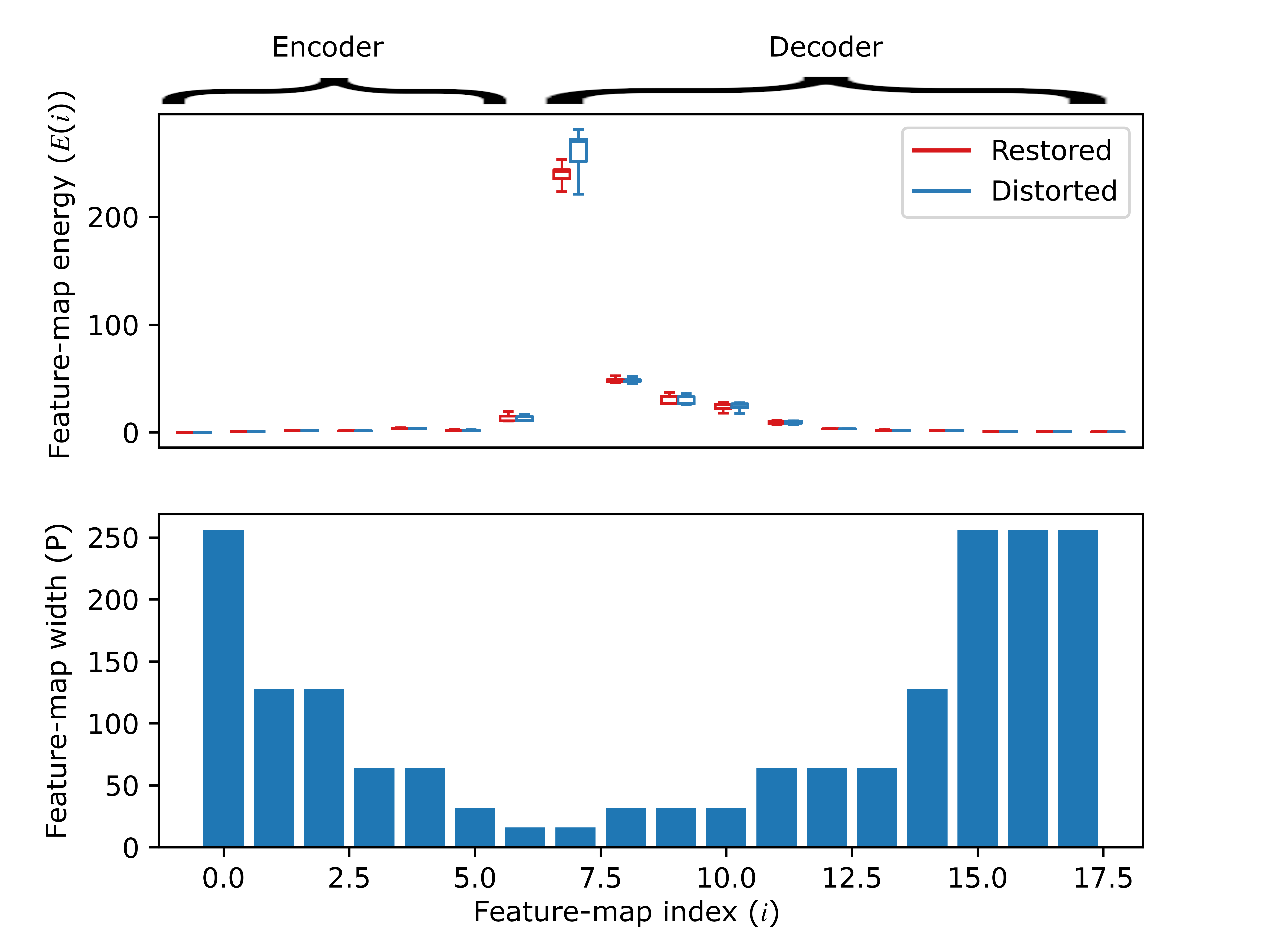}
         \caption{UNet-4 Levels}
         \label{fig:unet4}
     \end{subfigure}
     \hfill
     \begin{subfigure}[b]{0.31\textwidth}
         \centering
         \includegraphics[width=\textwidth]{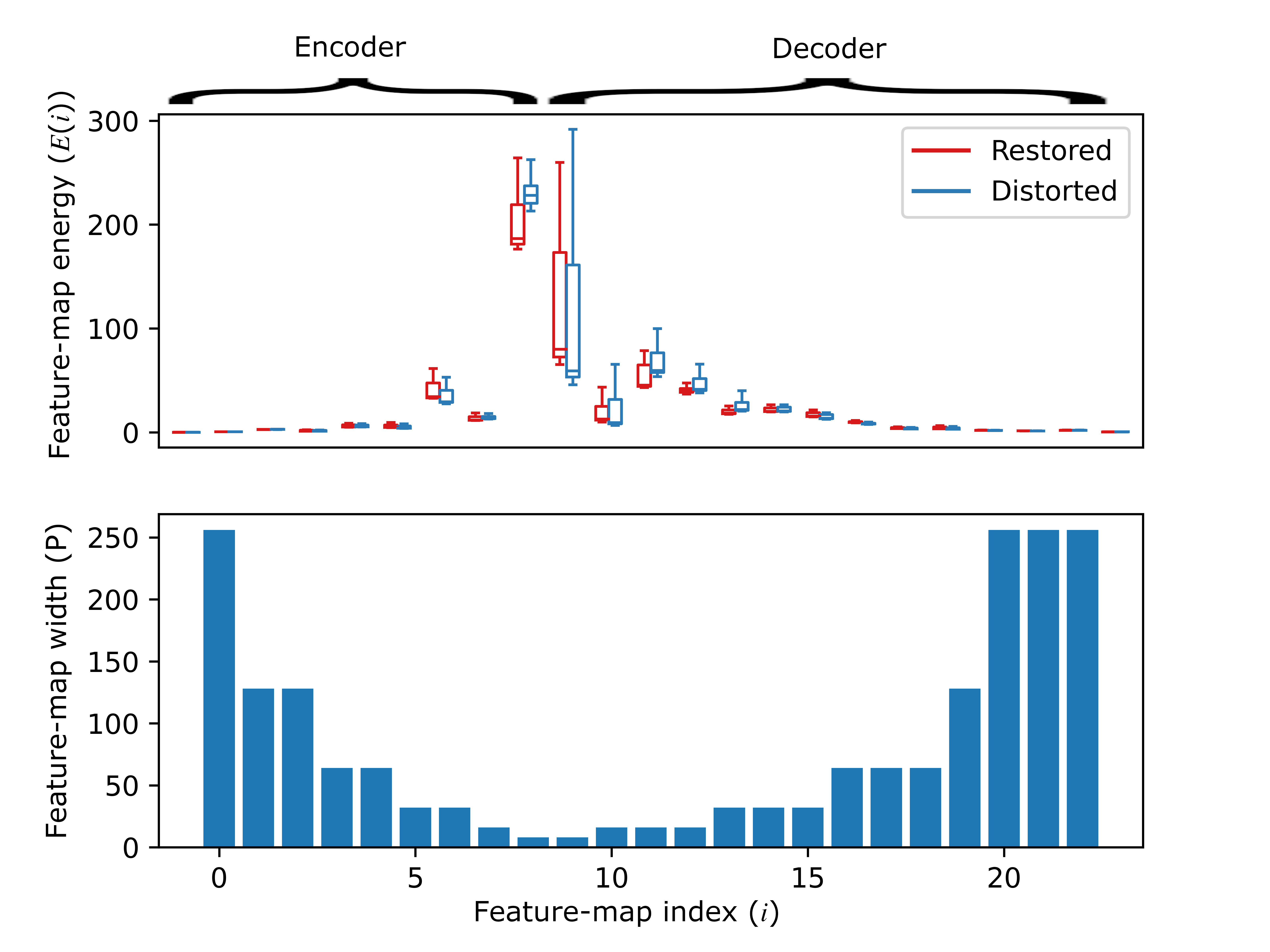}
         \caption{UNet-5 Levels}
         \label{fig:unet5}
     \end{subfigure}
      \caption{Feature-map analysis Van sequence: 3-4 levels ($\sim$200 frames, model: UNet)}
      \label{fig:unets34}
\end{figure}

\begin{figure}
     \centering
     \begin{subfigure}[t]{0.33\textwidth}
         \centering
         \includegraphics[width=\textwidth]{BigFigs/van_boxplot_unet5.png}
         \caption{Van sequence: UNet 5 levels.  PSNR: 27.0814, SSIM: 0.8365}
         \label{fig:unet5_normal}
     \end{subfigure}
     \begin{subfigure}[t]{0.33\textwidth}
         \centering
         \includegraphics[width=\textwidth]{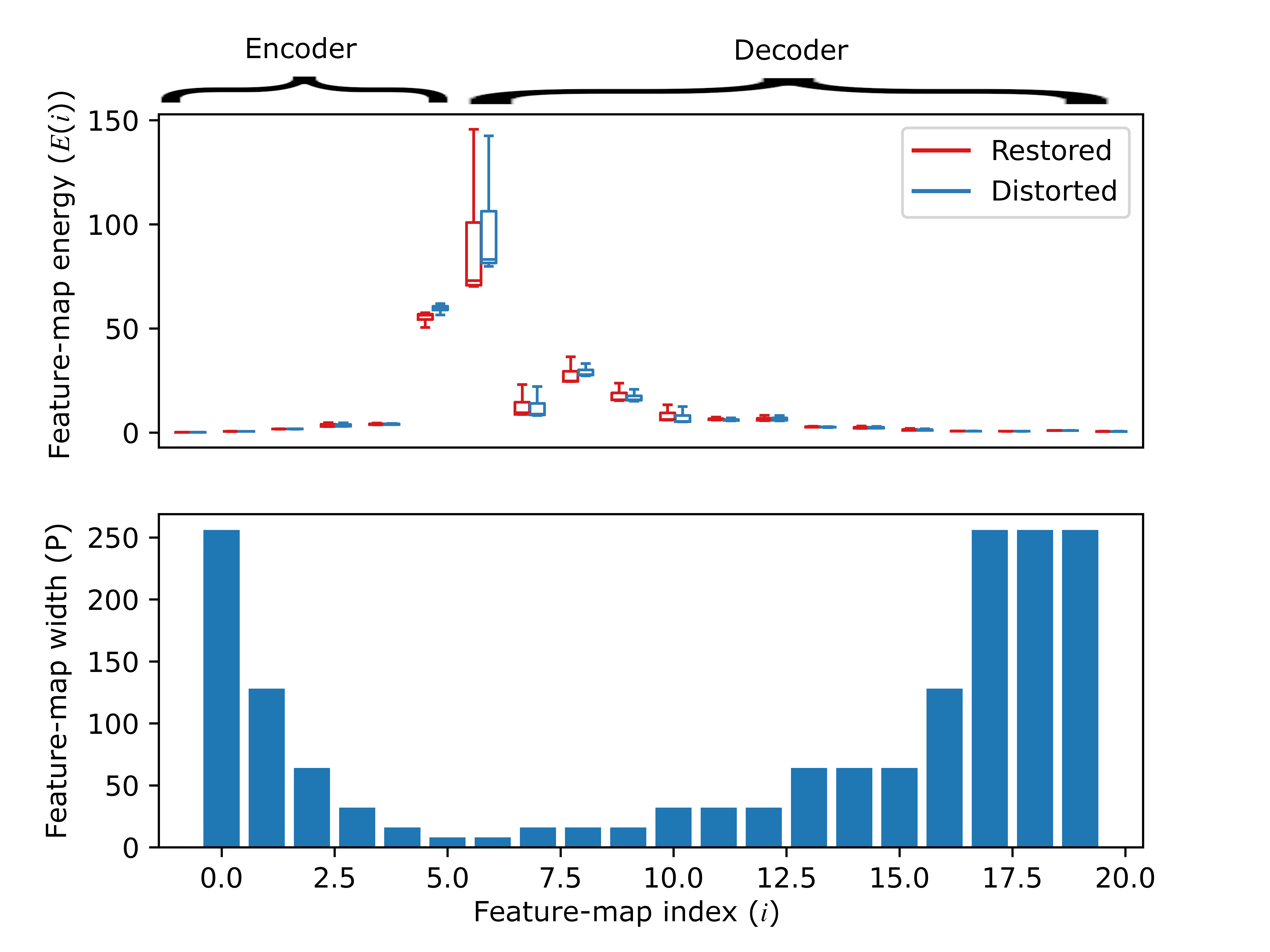}
         \caption{Van sequence: UNet 5 levels (with reduced encoder levels). PSNR: 27.3180, SSIM: 0.8332}
         \label{fig:unet5Red}
     \end{subfigure}
      \caption{Encoder level analysis}
      \label{fig:unetsRed}
\end{figure}

Figures \ref{fig:unet3}-\ref{fig:unet5}  show a range of results for the Van sequence trained on the remaining 11 other sequences.  Firstly, a UNet model is trained and the UNet depth is varied from 3 to 5. These plots show that there is very little ``energy'' in the encoder levels of the analysed UNet, implying less significance for the atmospheric turbulence mitigation task.  We therefore conduct a further analysis to reduce the number of convolutional layers in the encoder side to see the effect on performance.  Figure \ref{fig:unet5Red} illustrates the performance of a modified UNet architecture with the repeated Conv2D layers removed in the encoder (a standard UNet architecture is shown in Figure \ref{fig:unet5_normal}).  It was anticipated that due to the lack of energy within the encoder feature maps a large number of convolutional layers are not needed.  Initial estimates for PSNR and SSIM (for just one entire sequence) indicate that this is the case (they do not vary significantly when the repeated convolution layers are removed in the encoder).  Furthermore, this shows that the removal of redundant layers may not affect performance but will enable a much faster implementation (together with a reduced memory footprint).

The above analysis shows that across all sequences and network architectures, the encoder network contains less energy reflecting the blurred nature of input sequences (and therefore the smaller response of the ``early vision'' type filters learned in the early convolutional layers).  Furthermore, the deeper layers containing more abstract and semantic information at the centre of the encoder-decoder structure show more energy responses. This illustrates that the spatially higher resolution layers, which capture low-level features, are not being used for degradation mitigation compared to those layers capturing more semantic content and therefore able to reconstruct a more faithful result.
However, it is not certain that such a deduction would necessarily hold true for all of the wide range of the considered network architectures (as discussed in section~\ref{sec:deep}) for the next generation of atmospheric mitigation network-based solutions. 
\section{Conclusion}
\label{sec:conclusion}

This paper primarily gives an overview of the historical and state-of-the-art methods for turbulence mitigation. Additionally, the characteristics and causes of atmospheric turbulence are discussed together with a review of datasets (real and artificial) with a further review of methods to create fully artificial atmospheric turbulence (often employed in deep learning mitigation methods).  Finally, an analysis of feature map energy was included whereby a large range of test dataset sequences and network architectures were analysed (in terms of the feature map energies).  Since the subject of turbulence mitigation and deep learning in general is a fast moving research area, many new types of model architectures (such as INR and ViT architectures) will need new analysis going forward.  This paper has therefore concentrated on reviewing and (where possible) trying new architectures together with an appropriate level of analysis of conventional CNN based architectures.  

\small
\bibliography{literature_review,sim_lit}
\end{document}